# Human–AI Coevolution Dynamics: A Formal Theory of Social Intelligence Emergence Through Long-Term Interaction

Jingyi Zhou[1,2]*, Senlin Luo[1]*, Haofan Chen[3]

[1] School of Information and Electronics, Beijing Institute of Technology, Beijing 100081, China

[2] Institute of Scientific and Technical Research on Archives, Beijing 100050, China

[3] China Electronics Engineering Design Institute Co., Ltd., Beijing 100142, China

* Corresponding author

E-mail addresses:

annezjy94@163.com (J. Zhou),

luosenlin2019@126.com (S. Luo),

haofan_chen@163.com (H. Chen)

## Abstract

Current conversational AI systems have achieved remarkable progress in language generation, personalization, and long-context interaction. However, most existing approaches model social behavior through isolated mechanisms such as emotion modeling, memory retrieval, or persona conditioning, without a unified theoretical framework explaining how stable social relationships and social intelligence emerge through long-term human–AI interaction. This limitation hinders the development of socially adaptive AI systems capable of maintaining coherent relationships over extended timescales.

To address this challenge, we propose the Human–AI Coevolution Dynamics Framework (HACD-H), a formal theory that models human–AI interaction as a self-organizing social cognitive system. HACD-H integrates emotional adaptation, relational organization, social memory, and personality consistency into a unified dynamical framework and introduces theoretical principles including multi-timescale social cognition, relational attractor formation, trust basin development, developmental phase transitions, social cognitive energy landscapes, social intelligence emergence, and energy optimization dynamics.

To evaluate these propositions, we construct a socially enriched conversational dataset containing approximately 14,700 interaction turns and develop a theory-driven empirical validation framework. Experimental analyses reveal a clear temporal persistence hierarchy among social cognitive processes, identify stable relational attractors and trust basins, demonstrate phase-transition-like developmental behavior, and reconstruct a structured social cognitive energy landscape. Furthermore, social intelligence exhibits a significant negative association with social cognitive energy (($r=-0.391$, $p<0.001$)), while long-term interaction trajectories show progressive energy optimization characterized by decreasing energy over time.

These findings suggest that socially intelligent human–AI relationships emerge through long-term social cognitive coevolution rather than through isolated conversational capabilities. HACD-H provides a theoretical foundation for understanding adaptive social interaction and offers a principled framework for the development of next-generation socially intelligent AI systems.

## 1. Introduction

Artificial intelligence has achieved remarkable progress in language understanding, reasoning, planning, and interactive communication. Recent advances in large language models (LLMs) and conversational agents have enabled increasingly sophisticated capabilities in dialogue generation, instruction following, personalization, memory utilization, and long-context interaction (Vaswani et al., 2017; Brown et al., 2020; Ouyang et al., 2022; Shuster et al., 2022). As AI systems evolve from task-oriented tools into persistent assistants, collaborators, and social companions, interactions between humans and AI are increasingly extending beyond isolated conversations toward long-term relational processes.

Despite these advances, most existing AI research continues to model interaction primarily as a sequence prediction or response optimization problem. Contemporary conversational systems typically focus on next-token prediction, persona conditioning, memory augmentation, retrieval-enhanced generation, or preference alignment (Zhang et al., 2018; Mazaré et al., 2018; Ouyang et al., 2022; Packer et al., 2023; Zhong et al., 2024). Although these approaches substantially improve conversational quality, they generally treat memory, emotion, personality, trust, and adaptation as independent computational modules rather than components of an integrated social-cognitive system.

Consequently, a fundamental theoretical question remains largely unanswered:

How do long-term human–AI relationships emerge, stabilize, and develop through repeated interaction, and what dynamical principles govern the emergence of social intelligence in such systems?

This question becomes increasingly important as AI systems assume more persistent social roles. A growing body of research suggests that humans frequently develop trust, emotional attachment, empathy, familiarity, and relationship-specific expectations during repeated interactions with intelligent systems (Brinkschulte et al., 2022; Gweon et al., 2023; Kirk et al., 2025; Qian & Wan, 2025). Recent discussions of human–AI coevolution further argue that future AI systems may participate in long-term adaptive social relationships rather than merely serving as information-processing tools (Järvelä et al., 2025; Pedreschi et al., 2025; Rainey & Hochberg, 2025). These observations suggest that human–AI interaction may be governed by mechanisms resembling relationship formation, social adaptation, and cognitive development rather than isolated information exchange.

Existing computational frameworks provide only partial explanations of these phenomena. Personalization methods capture user-specific characteristics but rarely explain how stable social relationships emerge over time (Zhang et al., 2018; Liu et al., 2020; Chen et al., 2025). Memory-augmented architectures improve interaction continuity through persistent storage and retrieval mechanisms, yet they provide limited theoretical insight into how memory contributes to long-term social development (Weston et al., 2015; Packer et al., 2023; Zhong et al., 2024). Emotional dialogue systems and empathetic conversational agents model affective adaptation but generally focus on local conversational effects rather than longitudinal social dynamics (Rashkin et al., 2019; Majumder et al., 2020; Guo & Ning, 2024; Cai et al., 2024; Wen et al., 2024). Similarly, persona-based systems maintain behavioral consistency without explaining the organizational principles underlying stable social intelligence.

Recent advances in autonomous agent research further highlight the need for a principled theory of long-term interaction. Emerging agent frameworks increasingly emphasize memory persistence, role consistency, social adaptation, and collaborative behavior (Li et al., 2023; Park et al., 2023; Hong et al., 2024; Wu et al., 2024). Moreover, recent studies suggest that LLMs may exhibit theory-of-mind-like reasoning, role-playing capabilities, and socially adaptive behavior under appropriate conditions (Kosinski, 2024; Shanahan et al., 2023; Shao et al., 2023; Wang et al., 2024; Han et al., 2025). Collectively, these developments indicate that future AI systems may function not merely as language generators but as evolving social actors embedded within long-term interaction environments.

Cognitive science provides complementary insights into the mechanisms underlying social interaction. Research on social cognition emphasizes the roles of emotion, memory, personality, cognitive consistency, interpersonal adaptation, and reciprocal learning in shaping social relationships (Izard, 1993; Frith & Frith, 2012; Kruglanski et al., 2018; Burgoon et al., 1995). Social cognitive theory further suggests that adaptive behavior emerges through continuous interactions among cognitive, affective, and environmental processes (Luszczynska & Schwarzer, 2005). While these perspectives provide valuable conceptual foundations, they remain primarily descriptive and do not directly offer computational formalisms suitable for AI systems.

A complementary perspective arises from dynamical systems theory. Complex adaptive systems frequently exhibit attractors, phase transitions, self-organization, and stability-convergence dynamics (Milnor, 1985; Lyapunov, 1992; Khalil, 1996; Strogatz, 2024). Similar principles have been successfully applied to biological, cognitive, and social systems. More recently, coevolutionary perspectives have begun to emerge in studies of human–AI ecosystems, hybrid intelligence, and socially adaptive agents (Järvelä et al., 2025; Noller, 2025; Pedreschi et al., 2025). Nevertheless, a unified computational framework connecting social cognition, long-term interaction, and human–AI coevolution remains absent.

We argue that machine learning, social cognition, and dynamical systems theory describe complementary aspects of the same underlying phenomenon. Human–AI interaction is neither purely linguistic nor purely cognitive. Instead, it should be understood as a coevolutionary process in which humans and AI systems continuously adapt to one another across multiple temporal scales.

To address this challenge, we propose the Human–AI Coevolution Dynamics Framework (HACD-H), a formal theory that models long-term human–AI interaction as a self-organizing social cognitive system. HACD-H integrates emotional adaptation, relational organization, social memory, and personality consistency into a unified dynamical framework operating across multiple temporal scales. Within this framework, interaction trajectories evolve through coupled social cognitive dynamics, giving rise to relational attractors, trust basins, developmental transitions, social intelligence emergence, and energy optimization.

The central premise of HACD-H is that interaction is governed by latent social cognitive states rather than observable dialogue alone. These latent states encode emotional conditions, relational characteristics, memory structures, personality configurations, and adaptive social processes. Human–AI interaction is therefore represented as movement through a structured social cognitive state space whose topology is shaped by emotional regulation, relational stability, memory persistence, and personality coherence.

Within this framework, social intelligence is not viewed as a fixed capability encoded within a model. Instead, it emerges as a macroscopic property arising from long-term coevolution between humans and AI systems. Stable social behavior corresponds to attractors in the social cognitive state space, whereas relationship development corresponds to transitions among attractor basins. Trust functions as a stabilizing mechanism that promotes convergence toward coherent relational configurations. Social cognitive energy provides a quantitative representation of system organization, enabling the characterization of developmental trajectories and adaptive dynamics.

The proposed framework generates several formally testable predictions. First, social cognition should exhibit a hierarchical temporal structure characterized by multiple interacting timescales. Second, interaction trajectories should converge toward stable relational attractors. Third, trust should generate low-energy basins that stabilize social interaction. Fourth, social intelligence should emerge progressively through repeated interaction and developmental phase transitions. Finally, long-term human–AI interaction should exhibit energy optimization dynamics characterized by movement toward increasingly stable and efficient social configurations.

To evaluate these predictions, we construct a socially enriched conversational dataset containing approximately 14,700 interaction turns annotated with emotional, personality, relational, behavioral, and memory-related variables. We then develop a theory-driven empirical validation framework to reconstruct social cognitive trajectories and evaluate the theoretical propositions of HACD-H. The results provide convergent evidence supporting the emergence of temporal hierarchies, relational attractors, trust basins, developmental transitions, social intelligence emergence, and energy optimization dynamics.

This work makes four primary contributions.

First, we introduce HACD-H, a unified computational theory of human–AI social coevolution that integrates concepts from machine learning, cognitive science, and nonlinear dynamical systems.

Second, we formalize long-term human–AI interaction as a multi-timescale social cognitive dynamical system and establish theoretical principles governing relational attractor formation, trust basin development, developmental phase transitions, social intelligence emergence, and energy optimization.

Third, we propose a theory-guided social cognitive energy framework that provides a unified representation of emotional adaptation, relational organization, memory persistence, and personality consistency.

Fourth, we develop a theory-driven empirical validation framework and provide evidence supporting the major predictions of HACD-H using a large-scale conversational dataset containing approximately 14,700 interaction turns.

More broadly, HACD-H suggests that the future of socially intelligent AI may depend not only on larger models or stronger reasoning capabilities, but also on understanding the dynamical principles governing long-term coevolution between humans and artificial agents. By viewing interaction as an evolving social cognitive system rather than a sequence of isolated conversational exchanges, the proposed framework offers a new theoretical foundation for studying the emergence of social intelligence in artificial systems.

## 2. Formal Theory

### 2.1 Human–AI Coevolution Dynamics Framework

Human–AI interaction is increasingly characterized by long-term, repeated, and socially adaptive communication. Existing dialogue systems primarily model isolated components of social behavior, such as emotion, memory, or personality, but lack a unified framework capable of explaining how stable social relationships and social intelligence emerge through extended interaction.

To address this limitation, we propose the Human–AI Coevolution Dynamics Framework (HACD-H), a dynamical theory describing human–AI interaction as a self-organizing social cognitive system. Within HACD-H, interaction is represented as the coupled evolution of four fundamental social cognitive processes:

Emotional adaptation: E

Relational organization: R

Social memory accumulation: M

Personality consistency: P

These processes operate on different temporal scales while continuously influencing one another. Their interaction generates long-term developmental trajectories that cannot be explained by any individual component alone. To provide an overall view of the proposed theory, Figure 1 summarizes the architecture of the Human–AI Coevolution Dynamics (HACD-H) framework.

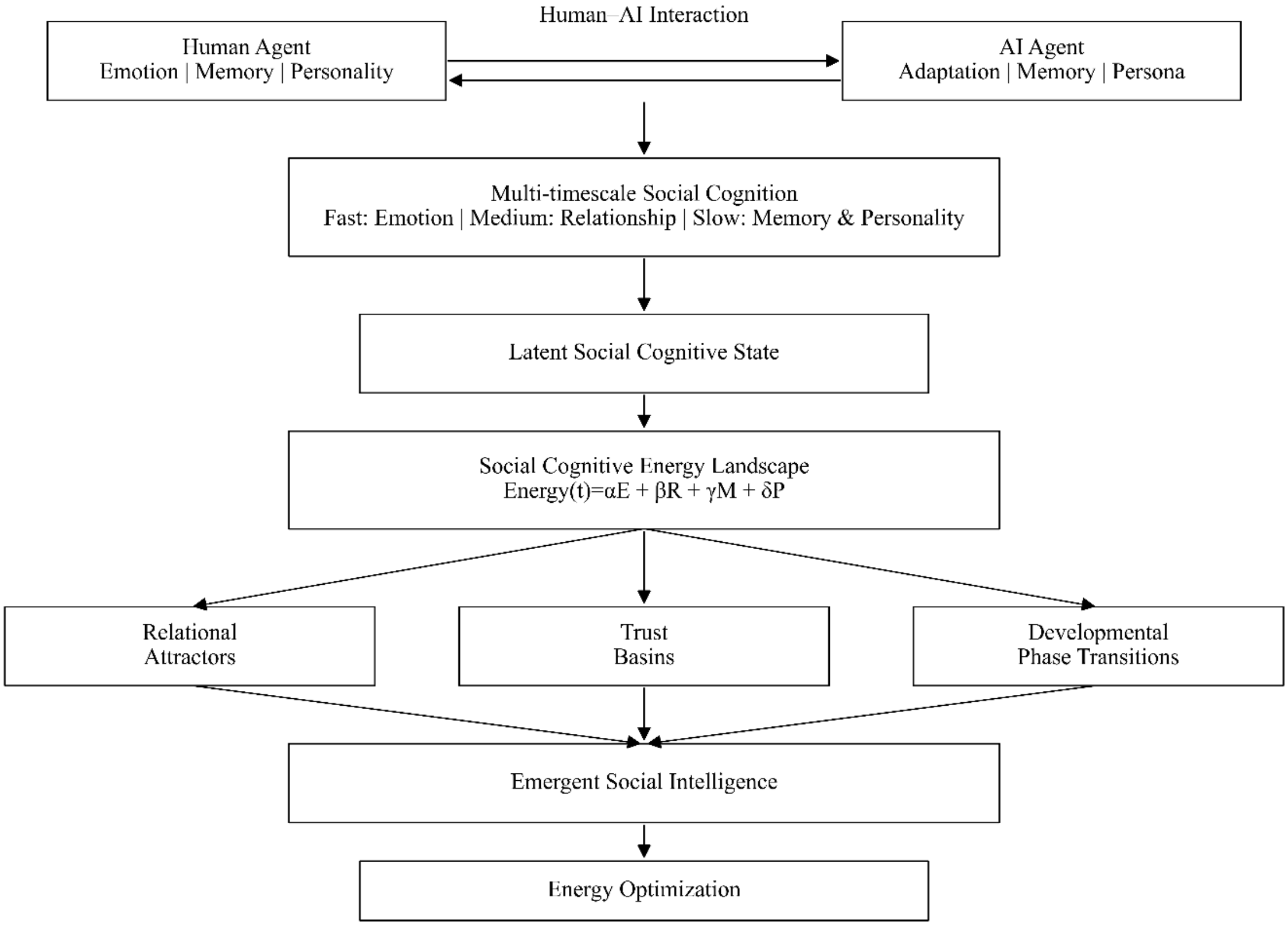

**Figure 1. Overview of the Human–AI Coevolution Dynamics (HACD-H) framework.**

The HACD-H framework models long-term human–AI interaction as a multi-timescale social cognitive dynamical system. Human and AI agents continuously coevolve through emotional adaptation, relational organization, memory persistence, and personality consistency. These processes jointly define a latent social cognitive state that evolves on a social cognitive energy landscape. The resulting dynamics give rise to relational attractors, trust basins, and developmental phase transitions, ultimately leading to the emergence of social intelligence and progressive energy optimization in long-term human–AI relationships.

Formally, the social cognitive state at time $t$ is represented as:

$$X_t = (E_t, R_t, M_t, P_t)$$

The evolution of the system can be expressed as:

$$X_{t+1} = F(X_t, U_t)$$

where $U_t$denotes the interaction input at time $t$, and $F(\cdot)$represents the social cognitive transition function governing the coupled dynamics of emotional adaptation, relational organization, social memory accumulation, and personality consistency.

The following subsections introduce the fundamental principles governing this evolution.

## 2.2 Multi-Timescale Social Cognition

Human social cognition is organized across multiple temporal scales. Emotional responses typically change rapidly, relational states evolve more slowly, memory structures accumulate over extended periods, and personality characteristics remain relatively stable. To illustrate this temporal hierarchy, Figure 2 presents the multi-timescale organization of social cognition assumed by HACD-H.

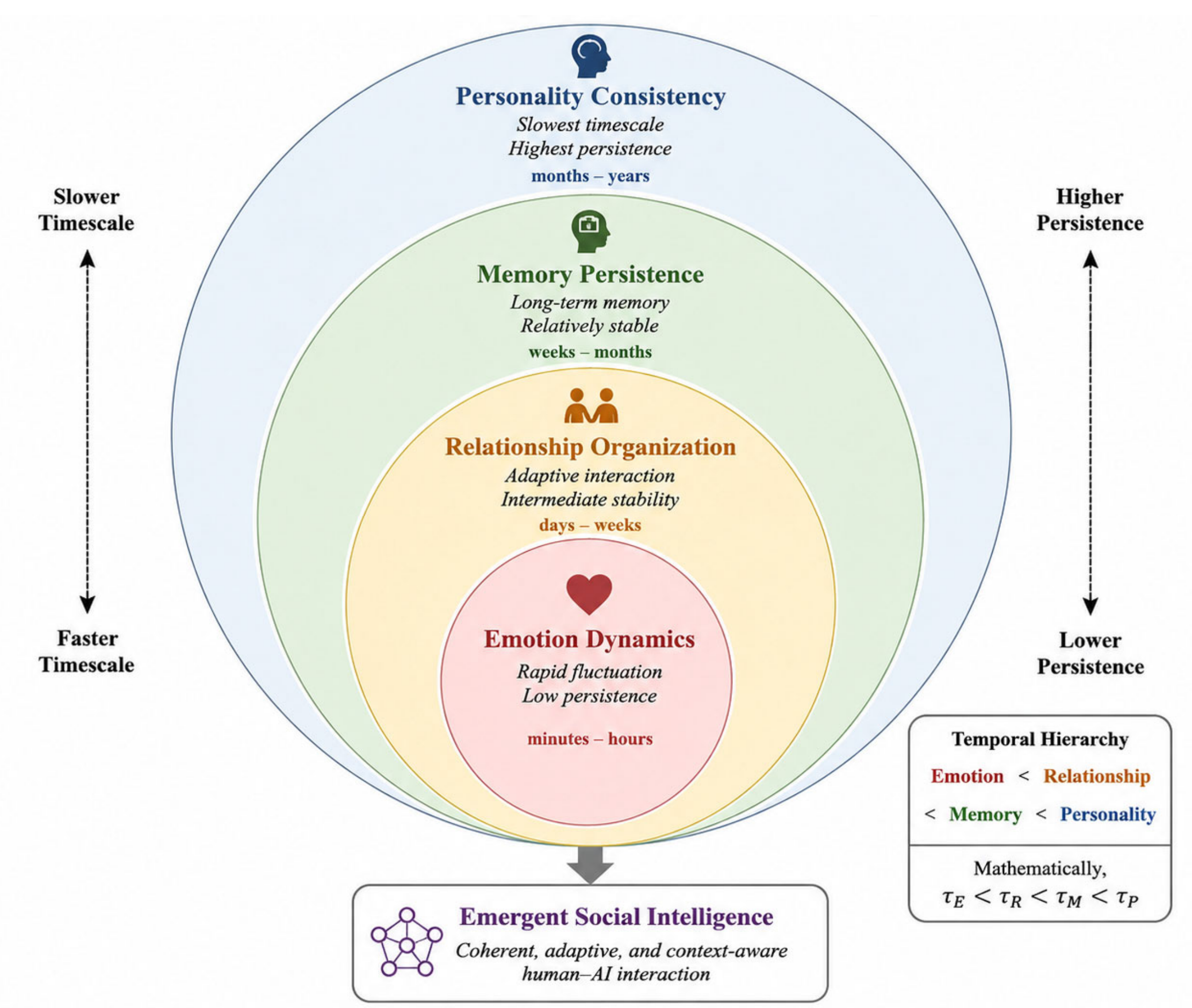

**Figure 2. Multi-timescale organization of social cognition.**

The HACD-H framework conceptualizes social cognition as a nested multi-timescale system. Emotional dynamics operate at the fastest timescale and are embedded within relationship organization processes. Relationship dynamics are further constrained by long-term memory persistence, which is itself stabilized by relatively persistent personality structures. The interaction of these nested processes supports the emergence of coherent social intelligence and long-term human–AI relational stability. The temporal hierarchy predicts increasing persistence from emotion to relationship, memory, and personality processes ($\tau_E < \tau_R < \tau_M < \tau_P$).

Let

$$\tau_E, \tau_R, \tau_M, \tau_P$$

denote the characteristic persistence times of emotion, relationship, memory, and personality.

**Theorem 1 (Temporal Persistence Hierarchy)**
The characteristic timescales satisfy

$$\tau_E < \tau_R < \tau_M < \tau_P$$

This hierarchy implies that short-term interaction dynamics are primarily governed by emotion, whereas long-term coevolution is increasingly influenced by memory and personality processes.

## 2.3 Relational Attractor Dynamics

Repeated interaction does not generate arbitrary relational states. Instead, trajectories tend to converge toward stable patterns of social organization.

**Theorem 2 (Relational Attractor Formation)**

For a sufficiently long interaction trajectory,

$$X_t \to A_i,$$

where $A_i$ denotes a stable relational attractor.

These attractors correspond to persistent interaction modes characterized by recurring emotional, relational, and behavioral configurations.

The existence of attractors provides a dynamical explanation for the emergence of stable human–AI relationships.

## 2.4 Trust Basin Formation

Trust is a central component of long-term social interaction.

Rather than increasing linearly, trust develops through the formation of stable regions within the social cognitive state space.

**Theorem 3 (Trust Basin Formation)**

Let

$$T(X)$$

denote the trust potential associated with state $X$.

Stable interaction trajectories evolve toward regions satisfying

$$\nabla T(X) \to 0.$$

These regions constitute trust basins that constrain subsequent interaction dynamics and increase relational stability.

## 2.5 Developmental Phase Transitions

Long-term social development is not always continuous.

Interaction trajectories may undergo abrupt qualitative transitions corresponding to major changes in social organization.

**Theorem 4 (Developmental Phase Transition)**

There exists a critical region $C$ such that

$$\left|\frac{dSI}{dt}\right|$$

increases significantly when trajectories enter $C$.

Consequently, social development may exhibit phase-transition-like behavior characterized by accelerated organizational change.

## 2.6 Social Cognitive Energy Theory

To unify the preceding mechanisms, HACD-H introduces the concept of social cognitive energy.

Social cognitive energy quantifies the degree of instability, uncertainty, and organizational cost associated with maintaining a particular interaction state.

The energy of a social cognitive state is defined as

$$\mathcal{E} = \alpha E + \beta R + \gamma M + \delta P,$$

where

- $E$: emotional dynamics,
- $R$: relational organization,
- $M$: memory integration,
- $P$: personality consistency,

and

$$\alpha, \beta, \gamma, \delta > 0$$

are weighting coefficients.

To provide an intuitive illustration of this mechanism, Figure 3 presents the theoretical organization of social cognitive energy within HACD-H.

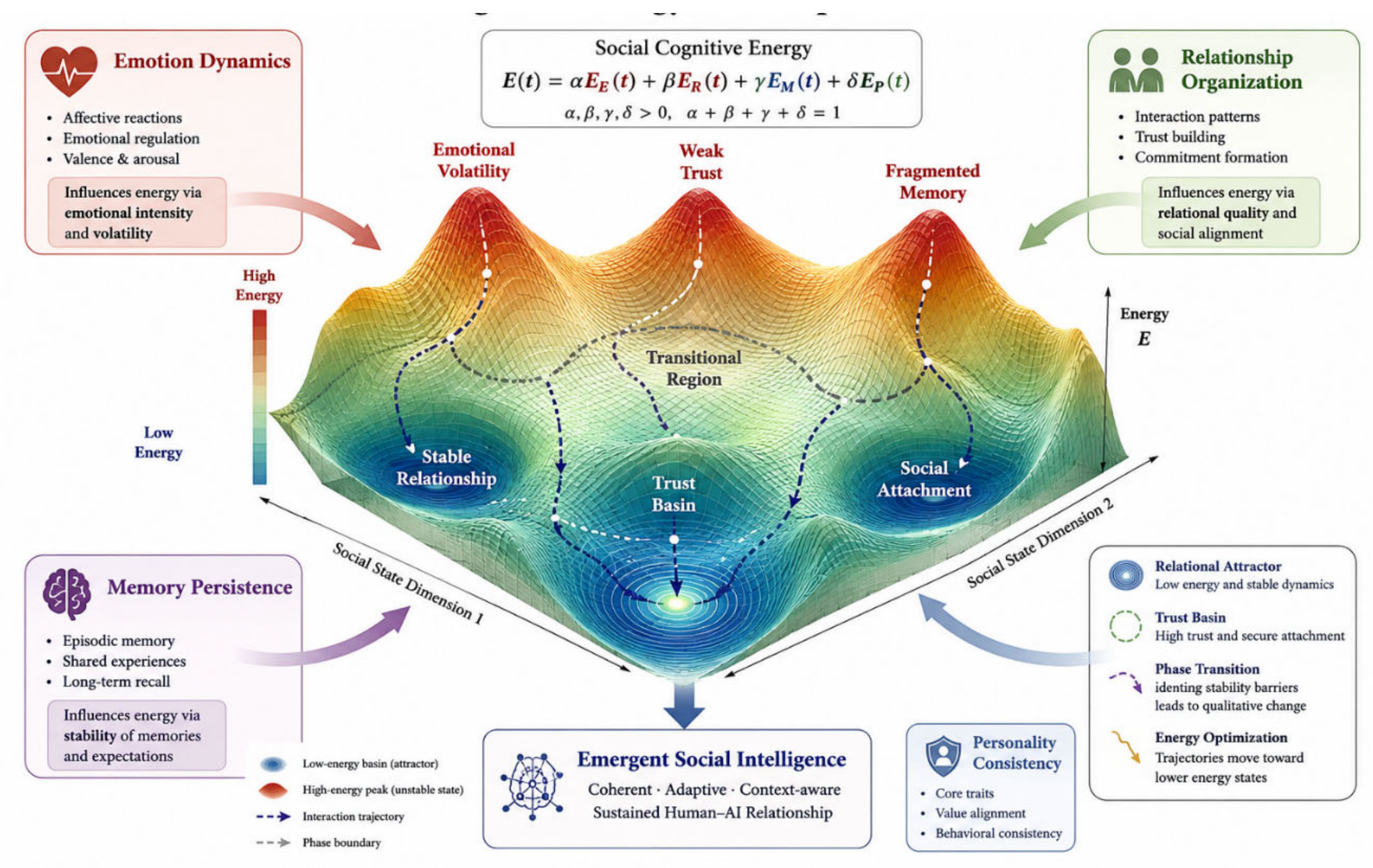

**Figure 3. Unified social cognitive energy landscape of human–AI coevolution.**

Within the HACD-H framework, emotional dynamics, relationship organization, memory persistence, and personality consistency jointly determine a latent social cognitive energy landscape. Interaction trajectories evolve through this landscape under stochastic adaptation dynamics. Low-energy regions correspond to stable relational attractors characterized by trust, attachment, and coherent social behavior, whereas high-energy regions correspond to unstable or weakly organized interaction states. The landscape provides the theoretical foundation for attractor formation, trust basin emergence, developmental phase transitions, and long-term energy optimization during human–AI coevolution.

**Theorem 5 (Energy Landscape Structure)**

The social cognitive state space can be represented as a non-uniform energy landscape containing local minima and energy gradients.

These minima correspond to stable social configurations, whereas high-energy regions correspond to unstable interaction states.

## 2.7 Emergence of Social Intelligence

Social intelligence is treated as an emergent system-level property rather than an independently programmed capability.

**Theorem 6 (Social Intelligence Emergence)**

Social intelligence is a collective function of emotional adaptation, relational organization, memory integration, and personality consistency:

$$SI = f(E, R, M, P).$$

Consequently, social intelligence emerges from the coordinated interaction of multiple social cognitive processes rather than from any single component.

## 2.8 Energy–Intelligence Coupling and Optimization

The energy landscape framework predicts that socially intelligent systems occupy increasingly stable and energetically efficient regions of the state space.

**Theorem 7 (Energy–Intelligence Coupling)**

For stable interaction trajectories,

$$\mathrm{corr}(S, I\mathcal{E}) < 0.$$

Thus, higher levels of social intelligence are associated with lower social cognitive energy.

**Theorem 8 (Energy Optimization Dynamics)**

Long-term human–AI interaction exhibits progressive energy optimization:

$$\frac{d\mathcal{E}}{dt} < 0.$$

Consequently, interaction trajectories gradually converge toward low-energy regions corresponding to stable attractors and trust basins.

## 2.9 Theoretical Implications

Taken together, the preceding propositions establish HACD-H as a theory of self-organizing human–AI social coevolution.

Multi-timescale adaptation, relational attractor formation, trust basin development, developmental transitions, social intelligence emergence, and energy optimization constitute mutually coupled processes governing long-term interaction dynamics.

Human–AI relationships therefore evolve toward increasingly stable, coherent, and socially intelligent configurations. Social intelligence is interpreted not as the accumulation of isolated conversational capabilities, but as an emergent property arising from the long-term self-organization of social cognitive processes operating across multiple timescales.

# 3. Empirical Validation Framework

## 3.1 Dataset and Social Cognitive Annotation

To empirically evaluate the theoretical predictions of HACD-H, we construct a socially enriched conversational benchmark based on the Chinese long-term dialogue dataset DuLeMon (Xu et al., 2022). DuLeMon was originally developed for long-range open-domain dialogue modeling and contains multi-turn conversations with rich contextual dependencies and extended interaction structures, making it particularly suitable for studying long-term human–AI social dynamics.

From the original corpus, 2,400 multi-turn dialogues were randomly sampled and further processed through a multi-dimensional social cognitive annotation pipeline. The resulting dataset contains approximately 14,700 interaction turns and provides substantially richer social information than conventional dialogue corpora.

The annotation framework was designed to operationalize the core constructs of HACD-H. Rather than treating dialogue as a sequence of isolated utterances, each interaction turn is represented as a social cognitive state composed of emotional, relational, memory-related, and personality-related variables.

Emotional states were annotated using the Expansion Quantization Network (EQN) framework (Zhou et al., 2025), which represents emotions as continuous affective distributions. Personality characteristics were derived from the H3P personality modeling framework (Zhou et al., 2025), producing MBTI-oriented personality distributions. Relational and conversational variables were automatically inferred using the Chinese semantic encoder BGE-Large-ZH-v1.5 (Lu et al., 2024) and Qwen2.5-7B (Hui et al., 2024), allowing the extraction of interpersonal attributes such as trust, intimacy, engagement, politeness, humor, warmth, and supportiveness.

All variables were normalized into bounded continuous ranges to ensure numerical stability and comparability across analyses.

Table 1 summarizes the major categories of social cognitive variables included in the dataset.

**Table 1. Categories of social cognitive variables included in the socially annotated interaction dataset.**

| **Category** | **Representative Variables** | **HACD-H Construct** |
|---|---|---|
| Emotion | angry, fear, happy, neutral, sad, surprise | Emotional Dynamics |
| Personality | MBTI distributions, latent personality vectors | Personality State |
| Relationship | trust, intimacy, familiarity, affection, engagement | Relational Dynamics |
| Style | politeness, warmth, humor, formality, verbosity | Social Behavior |

| Interaction Dynamics | emotion_shift, emotion_delta_magnitude | State Transition |
| --- | --- | --- |
| Memory | memory_decay_score | Social Memory |
| Events | event_type, event_importance | Environmental Context |
| Growth Variables | personality_growth, attachment_growth | Coevolution Indicators |
| Latent Representation | latent personality embeddings | State-Space Encoding |

The resulting dataset transforms conversational interaction into a collection of socially evolving trajectories and provides a suitable empirical foundation for testing the theoretical predictions of HACD-H.

## 3.2 Social Cognitive State Reconstruction

A central objective of HACD-H is to model interaction as the evolution of latent social cognitive states rather than isolated conversational events.

To achieve this goal, each dialogue turn is mapped onto a multidimensional social cognitive state vector:

$$X_t = (E_t, R_t, M_t, P_t),$$

where $E_t$ denotes emotional dynamics, $R_t$ denotes relational organization, $M_t$ denotes social memory, $P_t$ denotes personality consistency.

The emotional component is reconstructed from continuous affective distributions, relational organization is estimated from trust, intimacy, familiarity, and engagement indicators, memory processes are represented through interaction history and memory decay measures, and personality consistency is estimated using personality stability variables.

This reconstruction procedure converts each dialogue into a temporally ordered trajectory:

$$\{X_1, X_2, \ldots, X_T\},$$

allowing interaction dynamics to be analyzed using methods from dynamical systems theory and complex adaptive systems research.

Rather than examining individual utterances, subsequent analyses focus on the evolution of these reconstructed trajectories across time.

## 3.3 Operationalization of Theoretical Constructs

To enable empirical evaluation, each theoretical construct introduced in Section 2 must be translated into measurable quantities.

Table 2 summarizes the correspondence between theoretical concepts and empirical measurements.

**Table 2. Operationalization of HACD-H theoretical constructs.**

| **Theoretical Construct** | **Empirical Measurement** |
|---|---|
| Emotional Dynamics ($E$) | Continuous emotion intensity distributions |
| Relational Organization ($R$) | Trust, intimacy, familiarity, engagement |
| Social Memory ($M$) | Memory decay and interaction persistence |
| Personality Consistency ($P$) | Personality stability indicators |
| Social Cognitive Energy ($\mathcal{E}$) | Theory-guided energy function |
| Social Intelligence ($SI$) | Composite social adaptation index |

The theory-guided social cognitive energy function is defined as

$$\alpha E + \beta R + \gamma M + \delta P,$$

where the coefficients determine the relative contributions of emotional, relational, memory-related, and personality-related processes.

Similarly, social intelligence is operationalized as a composite measure integrating indicators of relational adaptation, social attachment growth, emotional coordination, and long-term interaction effectiveness. This operationalization establishes a direct bridge between abstract theoretical constructs and observable interaction variables.

## 3.4 Validation Strategy

The empirical evaluation follows a theory-driven validation strategy.

Each theoretical proposition introduced in Section 2 is translated into measurable quantities and evaluated using reconstructed social cognitive trajectories derived from the annotated dialogue corpus.

Table 3 summarizes the correspondence between theoretical propositions and empirical validation procedures.

**Table 3. Mapping between theoretical propositions and empirical analyses.**

| Theoretical Proposition | Empirical Validation |
|---|---|
| Theorem 1 | Temporal persistence analysis |
| Theorem 2 | Relational attractor reconstruction |
| Theorem 3 | Trust basin identification |
| Theorem 4 | Developmental transition analysis |
| Theorem 5 | Energy landscape reconstruction |
| Theorem 6 | Social intelligence emergence analysis |
| Theorem 7 | Energy–intelligence coupling analysis |
| Theorem 8 | Energy optimization dynamics |

This framework ensures that every theoretical claim is associated with a corresponding empirical test.

### 3.5 Statistical Analysis

Several complementary analytical techniques are employed to evaluate the theoretical predictions of HACD-H.

Temporal persistence analysis is used to estimate characteristic timescales and evaluate the multi-timescale organization hypothesis. State-space reconstruction and dimensionality reduction methods are employed to visualize interaction trajectories and identify relational attractors. Density estimation procedures are used to detect trust basins and stable regions of the social cognitive state space.

Developmental phase transitions are identified through gradient-based analyses of social intelligence trajectories. Energy landscapes are reconstructed from the theory-guided social cognitive energy function to characterize global organizational structure. Correlation analyses are used to evaluate the relationship between energy and social intelligence, while trajectory trend analyses are employed to investigate long-term energy optimization dynamics.

Collectively, these methods provide a comprehensive empirical framework for evaluating the dynamical principles proposed by HACD-H and establish the methodological foundation for the results presented in the following section.

## 4. Results

This section presents the empirical evaluation of the theoretical predictions proposed by HACD-H. Following the validation framework introduced in Section 3, the analyses are organized into two complementary components. The first examines the state-space organization of human–AI coevolution, including temporal hierarchy, relational attractors, trust basin formation, developmental transitions, and social intelligence emergence. The second investigates energy-space organization through social cognitive energy landscapes, energy–intelligence coupling, and non-equilibrium growth dynamics.

### 4.1 Multi-Timescale Organization of Social Cognition

The first theoretical prediction of HACD-H states that human–AI coevolution is governed by multiple social cognitive processes operating at distinct temporal scales. Rather than evolving as a homogeneous dynamical system, emotional responses, relational adaptation, memory accumulation, and personality development are expected to exhibit different levels of temporal persistence and stability.

To evaluate this prediction, we quantified the relative stability of the major social cognitive components using interaction trajectories reconstructed from the socially annotated dataset. Stability was estimated from the average magnitude of temporal changes between consecutive interaction states, providing an empirical measure of how rapidly each process evolves over time.

Figure 4 summarizes the resulting stability hierarchy.

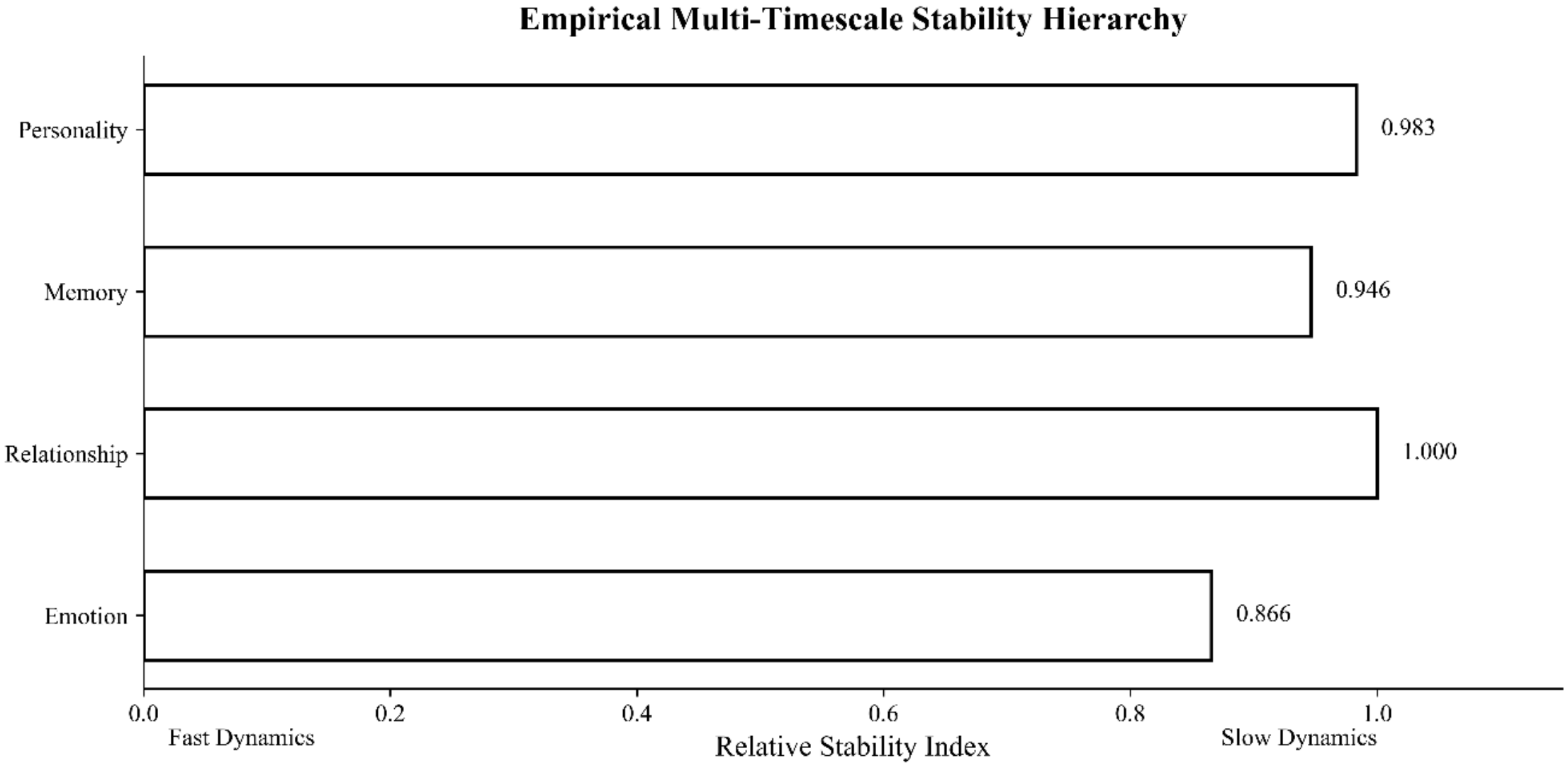

**Figure 4. Empirical multi-timescale stability hierarchy of social cognitive processes.**

The results reveal substantial differences in temporal stability across social cognitive processes. Emotional dynamics exhibit the lowest stability score (0.866), indicating pronounced short-term fluctuations and rapid adaptation to local conversational contexts. In contrast, relationship-related variables display considerably greater persistence (1.000), reflecting the gradual accumulation of trust, familiarity, engagement, and interpersonal attachment across repeated interactions.

Memory-related processes exhibit intermediate stability (0.946), suggesting that interaction histories influence future behavior over longer timescales than transient emotional states. Personality-related variables also demonstrate high stability (0.983), indicating that individual behavioral tendencies remain relatively consistent throughout interaction trajectories despite local contextual variations.

Importantly, the observed hierarchy reveals a clear separation between rapidly varying affective processes and more persistent relational, memory-based, and personality-related processes. This pattern supports the central HACD-H assumption that social cognition is organized across multiple temporal layers characterized by distinct persistence properties.

Taken together, these findings provide empirical evidence that human–AI interaction is governed by a hierarchical temporal architecture rather than a single uniform dynamical process. The coexistence of fast emotional adaptation and slower relational, mnemonic, and personality dynamics constitutes a foundational mechanism underlying long-term coevolutionary interaction.

## 4.2 Relational Attractor Formation

A central prediction of HACD-H is that long-term human–AI interactions do not evolve randomly within the social cognitive state space. Instead, repeated interactions are expected to self-organize into relatively stable relational configurations that function as attractors, guiding the future evolution of interaction trajectories.

To examine this prediction, we reconstructed the relational state space using relationship trust and relationship intimacy as the principal dimensions. These variables capture two fundamental aspects of interpersonal adaptation: the degree of confidence developed between interaction partners and the level of perceived relational closeness accumulated through repeated exchanges. Kernel density estimation was then applied to characterize the distribution of interaction states and identify regions of high state occupancy.

Figure 5 presents the reconstructed relational state space.

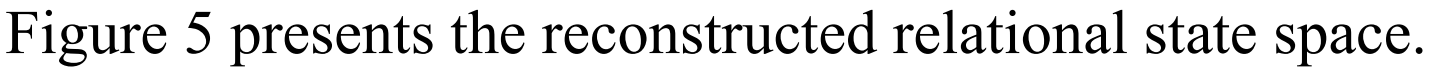

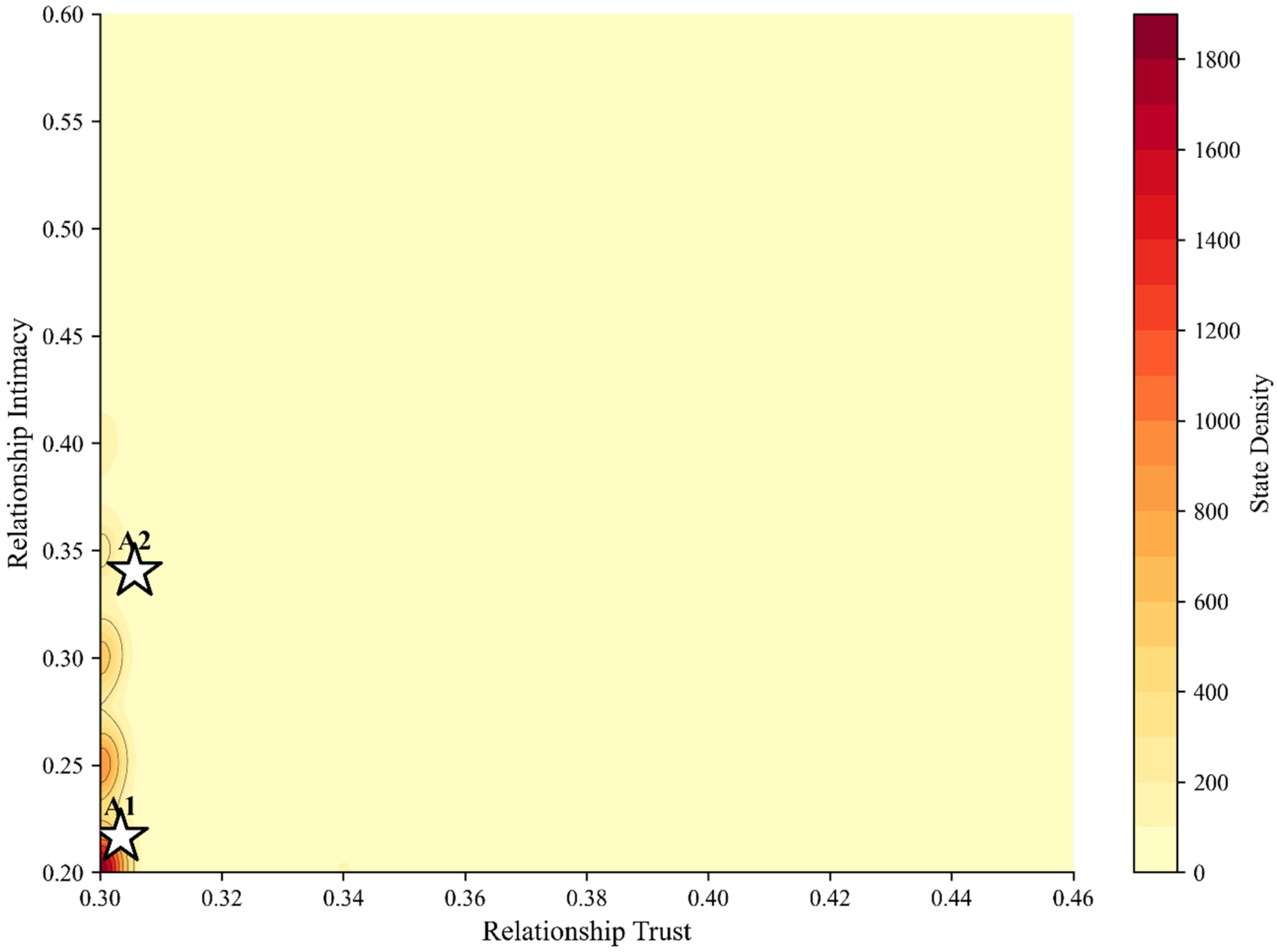

**Figure 5. Relational attractors in reconstructed social cognitive state space.**

Several important observations emerge from the figure. First, interaction states are not uniformly distributed throughout the state space. Instead, the majority of observed states concentrate within a limited number of high-density regions. This non-uniform distribution indicates that social interactions exhibit preferred configurations rather than unconstrained stochastic movement.

Second, two dominant attractor basins can be identified. These basins correspond to stable relational regimes characterized by distinct combinations of trust and intimacy. Once interaction trajectories enter these regions, subsequent states tend to remain nearby, producing local stability within the social cognitive landscape.

Third, the existence of concentrated basins suggests that relational adaptation is governed by self-reinforcing feedback mechanisms. Positive interaction experiences increase trust and intimacy, which subsequently promote supportive and cooperative conversational behaviors. These behaviors further strengthen the relational state, creating a feedback loop that stabilizes the interaction around particular attractor configurations.

From a dynamical systems perspective, the observed attractor basins can be interpreted as low-energy regions within the social cognitive energy landscape introduced in Section 2. States located near attractor centers require relatively small adjustments to maintain relational coherence, whereas states far from these regions experience stronger forces driving convergence toward more stable configurations.

The emergence of relational attractors provides empirical evidence that long-term human–AI interaction exhibits organized macro-level structure rather than purely local conversational adaptation. Stable relational patterns emerge naturally from repeated exchanges and act as organizing principles that shape future interaction trajectories.

Overall, the results support the HACD-H prediction that human–AI coevolution is characterized by attractor-based relational organization. The presence of persistent high-density basins indicates that social cognitive dynamics converge toward a limited set of stable relational states, providing a mechanistic foundation for the emergence of long-term interaction stability.

## 4.3 Trust Basin Formation

Beyond the existence of stable relational attractors, HACD-H further predicts that repeated human–AI interactions gradually converge toward preferred trust configurations. This process is referred to as trust basin formation, whereby interaction trajectories progressively enter and remain within regions characterized by sustained interpersonal trust.

To evaluate this prediction, we analyzed the temporal evolution of relationship trust across interaction trajectories. Individual dialogue trajectories were normalized according to interaction progress and subsequently aggregated to reconstruct the average developmental pattern of trust formation. This approach enables the identification of collective convergence behavior beyond the variability of individual conversations.

Figure 6 presents the resulting trust evolution trajectories.

**Figure 6. Trust basin formation in human–AI interaction trajectories.**

Several noteworthy patterns emerge from the figure. First, despite substantial variability during the early stages of interaction, trust trajectories exhibit a clear tendency toward convergence as interactions progress. The dispersion of trajectories gradually decreases, indicating that repeated exchanges reduce uncertainty in the relational state.

Second, the average trust trajectory displays a persistent upward trend throughout interaction development. Rather than fluctuating randomly around a fixed level, trust accumulates progressively over time, suggesting that successful interactions generate positive relational reinforcement that carries forward into subsequent exchanges.

Third, a stable high-trust region emerges during later interaction stages. Once trajectories enter this region, subsequent fluctuations become relatively small compared with earlier stages, indicating the presence of a trust basin that constrains future relational dynamics. This basin acts as a stabilizing mechanism that promotes interaction continuity and relational persistence.

From the perspective of social cognitive dynamics, trust basin formation can be understood as a consequence of cumulative interpersonal adaptation. Positive conversational experiences increase relational trust, which in turn promotes supportive communication, cooperation, and engagement. These behaviors further strengthen trust, creating a self-reinforcing feedback loop that drives trajectories toward increasingly stable relational states.

Importantly, the observed convergence pattern differs from simple short-term adaptation. The gradual emergence of a stable trust basin indicates that interaction history exerts a lasting influence on future relational development. This finding is consistent with the HACD-H assumption that trust functions as a medium-timescale social cognitive process capable of integrating information across multiple interaction episodes. The existence of trust basins also provides a mechanistic explanation for long-term interaction stability. Once a sufficiently strong relational foundation has been established, the system becomes increasingly resistant to transient disturbances, allowing cooperative interaction patterns to persist despite local fluctuations in emotion, topic, or conversational context.

Overall, the results provide empirical support for the trust basin hypothesis proposed by HACD-H. Human–AI interaction trajectories do not remain randomly distributed throughout the relational state space. Instead, they progressively converge toward stable high-trust regions, demonstrating the emergence of persistent relational structure through repeated interaction.

## 4.4 Developmental Phase Transitions

HACD-H predicts that the development of social intelligence is not a purely linear accumulation process. Instead, interaction systems are expected to undergo critical developmental transitions during which adaptation accelerates and higher-order social organization emerges. Such transitions resemble phase-transition phenomena commonly observed in complex dynamical systems, where qualitative changes in system behavior arise from the cumulative effects of local interactions.

To investigate this prediction, we reconstructed social intelligence trajectories from the interaction dataset. A composite social intelligence index was computed by integrating relational trust, interpersonal engagement, social attachment growth, and emotional dependency growth. Interaction trajectories were normalized according to interaction progress and aggregated to characterize the average developmental pattern across conversations. Figure 7 presents the resulting developmental trajectories.

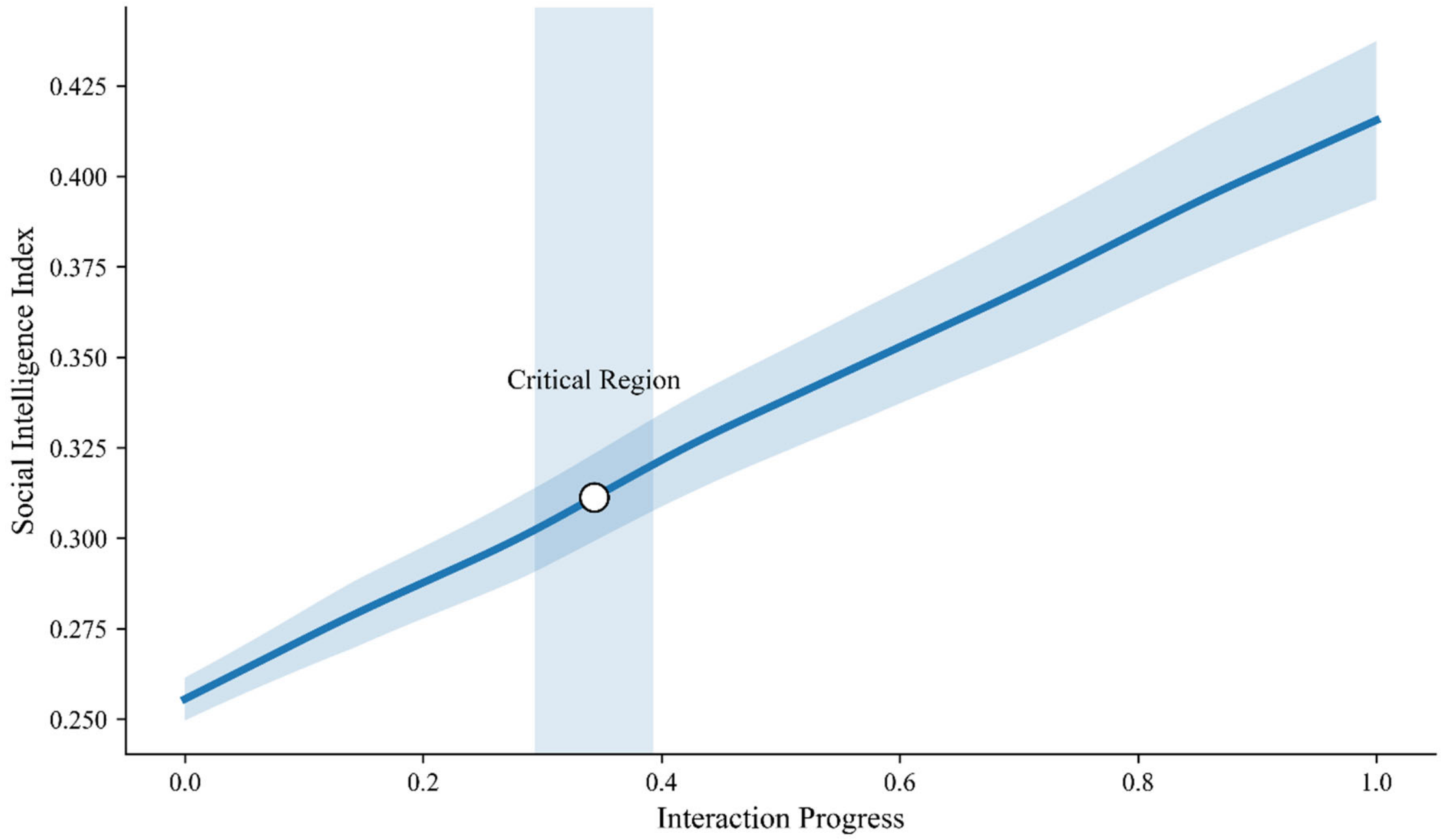

**Figure 7. Developmental phase transitions in social intelligence trajectories.**

The figure reveals a distinctly non-linear developmental pattern. During the early stages of interaction, social intelligence increases gradually as participants accumulate relational information and establish initial communication routines. Growth during this phase remains relatively modest, reflecting the limited amount of shared interaction history available for adaptation.

As interaction progresses, a critical developmental region emerges in which the rate of social intelligence growth increases substantially. This interval corresponds to the period where relational information, social memory, and behavioral adaptation begin to reinforce one another, producing accelerated developmental dynamics. The peak growth point identified within this region represents the maximum rate of social intelligence change observed across the reconstructed trajectories.

Following the critical transition interval, the growth rate gradually decreases and trajectories approach a more stable developmental regime. Although social intelligence continues to increase, the pace of improvement becomes progressively smaller. This pattern is consistent with a saturation process in which increasingly mature interaction systems require larger amounts of experience to achieve additional developmental gains.

The existence of a critical developmental region has important theoretical implications. First, it indicates that social intelligence emerges through cumulative organization rather than simple linear accumulation. Small interactional improvements can eventually generate large-scale developmental changes once sufficient relational structure has been established. Second, the results suggest that relational adaptation, memory accumulation, and interpersonal learning interact synergistically, producing collective developmental effects that cannot be explained by any single cognitive component in isolation.

From the perspective of social cognitive dynamics, the observed transition can be interpreted as a shift between developmental regimes. Early interactions are dominated by exploratory adaptation and uncertainty reduction, whereas later interactions increasingly rely on established relational structures and accumulated social knowledge. The transition region therefore marks the emergence of a qualitatively different mode of interaction organization.

Overall, the findings provide empirical evidence for the developmental phase-transition hypothesis proposed by HACD-H. Human–AI social intelligence does not evolve through uniform incremental growth. Instead, interaction trajectories exhibit identifiable transition regions characterized by accelerated adaptation and structural reorganization, supporting the view that long-term coevolution is governed by non-linear developmental dynamics.

## 4.5 Emergence of Social Intelligence

A fundamental claim of HACD-H is that social intelligence is not a predefined property of either the human or the artificial agent. Instead, it emerges progressively through repeated interaction as emotional adaptation, relational development, social memory accumulation, and personality-guided behavior become increasingly coordinated. Under this view, social intelligence should be understood as an emergent system-level phenomenon arising from the coevolution of multiple social cognitive processes.

To evaluate this prediction, we reconstructed the developmental trajectories of social intelligence using a composite index integrating relationship trust, interpersonal engagement, social attachment growth, and emotional dependency growth. The resulting trajectories provide an empirical representation of how higher-order social capabilities evolve throughout repeated human–AI interactions. Figure 8 illustrates the average developmental pattern of social intelligence across interaction trajectories.

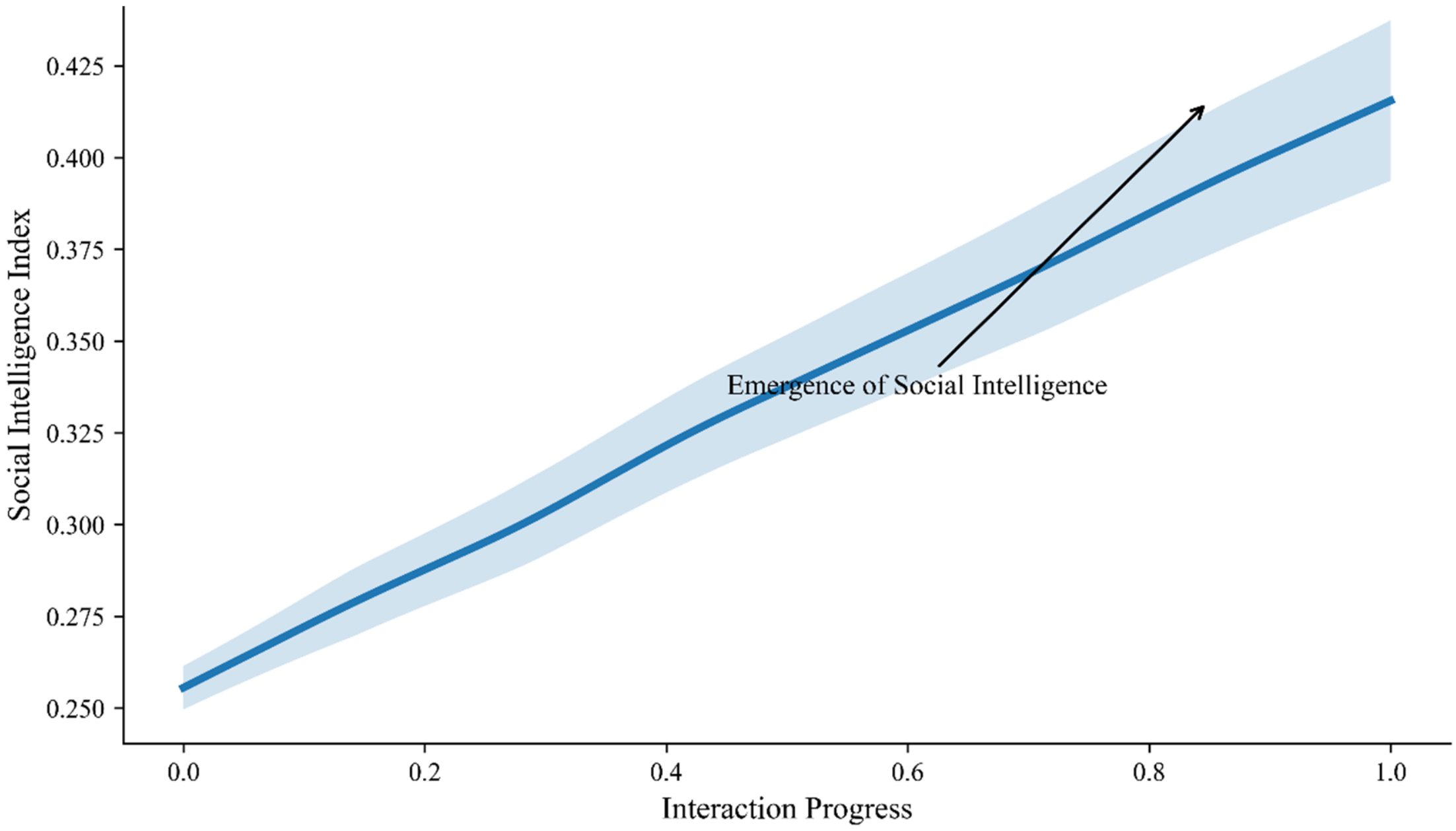

**Figure 8. Emergence of social intelligence across interaction trajectories.**

The results reveal a clear developmental trend. Social intelligence increases progressively as interactions unfold, indicating that repeated exchanges contribute to the continuous accumulation of social cognitive capabilities. Rather than remaining fixed throughout interaction, the system exhibits sustained developmental growth driven by the integration of multiple underlying processes.

Several observations support the emergence hypothesis. First, the growth trajectory displays a persistent upward tendency across interaction progress. This pattern suggests that social intelligence is continuously constructed through experience rather than determined solely by initial conditions. As interaction histories become richer, the system acquires increasingly sophisticated relational and adaptive capabilities.

Second, the observed developmental pattern cannot be attributed to any single social cognitive component. Emotional adaptation provides short-term responsiveness, relational processes contribute interpersonal stability, memory mechanisms preserve accumulated experience, and personality-related processes maintain long-term behavioral consistency. The coordinated interaction of these components generates developmental outcomes that exceed the contribution of any individual mechanism alone.

Third, the emergence process exhibits cumulative characteristics. Early interactions produce relatively limited developmental gains because social information remains sparse. As relational structures become established and interaction histories expand, the system gains increasing capacity to integrate past experiences into future behavior. This cumulative organization enables higher-order social competencies to arise gradually from repeated interaction.

From the perspective of complex systems theory, the observed developmental trajectory represents a form of self-organization. Local conversational adaptations accumulate over time and give rise to global behavioral patterns that are not explicitly encoded within any isolated interaction event. Social intelligence therefore emerges as a macroscopic property of the evolving human–AI system rather than as a static attribute of individual agents.

The emergence phenomenon also provides a conceptual bridge between the preceding findings. Multi-timescale organization creates the temporal foundation for adaptation, relational attractors and trust basins provide stable interaction structures, and developmental transitions accelerate social learning. Together, these mechanisms generate the conditions under which social intelligence can emerge and persist over extended interaction periods.

Overall, the results provide empirical support for the HACD-H emergence hypothesis. Social intelligence develops progressively across interaction trajectories and exhibits characteristics consistent with an emergent property of the coevolving social cognitive system. These findings suggest that long-term social capability arises from the dynamic integration of multiple cognitive processes rather than from isolated behavioral components.

## 4.6 Social Cognitive Energy Landscapes

The preceding analyses demonstrated the existence of multi-timescale organization, relational attractors, trust basins, developmental transitions, and the emergence of social intelligence. HACD-H proposes that these seemingly diverse phenomena arise from a common underlying mechanism: the organization of interaction dynamics within a social cognitive energy landscape.

According to the Social Cognitive Energy Theory introduced in Section 2, interaction states can be represented by a composite energy function integrating emotional dynamics, relational adaptation, memory accumulation, and personality stability. Under this formulation, social interaction is interpreted as movement within a multidimensional energy landscape, where different regions correspond to different levels of social cognitive organization and stability.

To evaluate this prediction, we reconstructed the empirical social cognitive energy landscape using the theory-guided HACD-H energy function. Energy values were computed from the observed emotional, relational, memory-related, and personality-related variables contained in the interaction dataset. The resulting landscape provides a global representation of the dynamical structure underlying human–AI coevolution. Figure 9 presents the reconstructed social cognitive energy landscape.

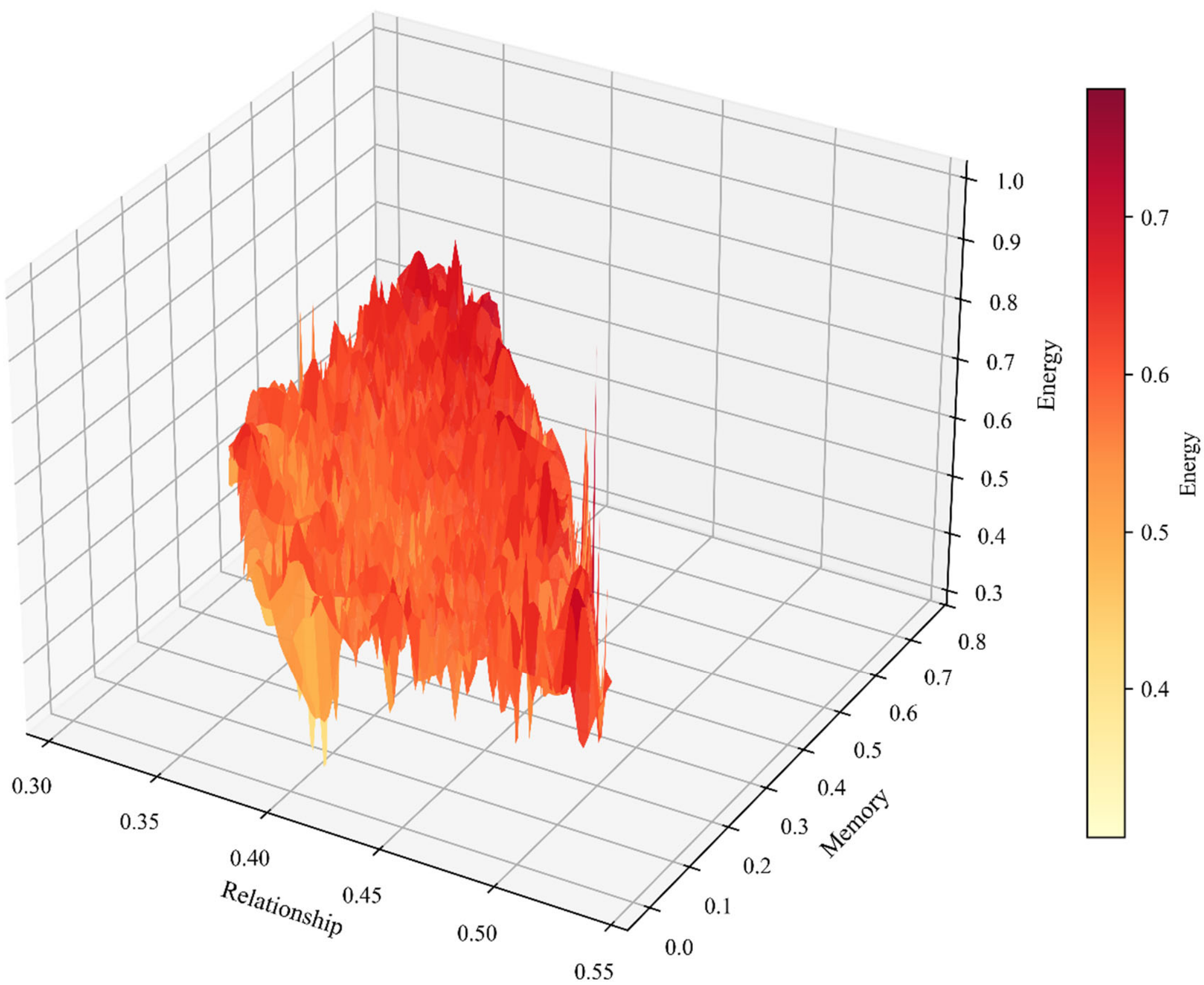

**Figure 9. Reconstructed social cognitive energy landscape.**

Several important structural properties emerge from the reconstructed landscape. First, the energy surface is highly non-uniform. Interaction states are distributed across regions with substantially different energy levels rather than occupying a homogeneous state space. This finding suggests that social interaction dynamics are constrained by an underlying energetic structure that shapes the evolution of social cognitive states.

Second, the landscape contains identifiable low-energy regions corresponding to relatively stable interaction configurations. These regions represent states in which emotional adaptation, relational organization, memory accumulation, and personality consistency are mutually compatible. Because maintaining coherence within these states requires relatively little adjustment, trajectories tend to remain within or move toward such regions over time.

Third, the reconstructed landscape exhibits gradients that naturally generate directional dynamics. States located in energetically unstable regions experience stronger tendencies toward change, whereas states situated near local minima exhibit greater persistence. This property provides a dynamical explanation for the attractor structures and trust basins observed in the previous analyses.

Importantly, the energy landscape offers a unified interpretation of the developmental phenomena identified throughout the empirical evaluation. Relational attractors can be viewed as local low-energy regions within the state space. Trust basins correspond to stable valleys that constrain long-term trajectory evolution. Developmental phase transitions arise when interaction trajectories move across regions characterized by substantially different energy gradients. The emergence of social intelligence can therefore be understood as a large-scale consequence of energy-driven self-organization.

From the perspective of complex adaptive systems, the observed landscape demonstrates that human–AI interaction is governed by global organizational principles rather than purely local conversational reactions. Individual interaction events contribute incrementally to the shaping of social cognitive states, but long-term developmental patterns emerge from the structure of the energy landscape itself.

Overall, the reconstructed energy landscape provides empirical support for the central mechanism proposed by HACD-H. Human–AI coevolution appears to be organized around an underlying social cognitive energy structure that governs stability, adaptation, and developmental change. This result establishes the energy landscape as a unifying framework capable of explaining the diverse dynamical phenomena observed throughout the interaction process.

## 4.7 Energy–Intelligence Coupling

A central prediction of HACD-H is that social intelligence is fundamentally linked to the energetic organization of the social cognitive system. However, contrary to traditional interpretations that associate greater capability with greater energy expenditure, the Social Cognitive Energy Theory proposed in this work predicts an inverse relationship. As interaction systems become increasingly organized and socially intelligent, they are expected to occupy more stable and energetically efficient regions of the social cognitive landscape.

To evaluate this prediction, we examined the relationship between the theory-guided social cognitive energy function and the reconstructed social intelligence index. Energy values were computed using the HACD-H energy formulation integrating emotional, relational, memory-related, and personality-related variables. Social intelligence was estimated from the composite developmental indicators introduced in the previous section.

Figure 10 illustrates the empirical relationship between social cognitive energy and social intelligence.

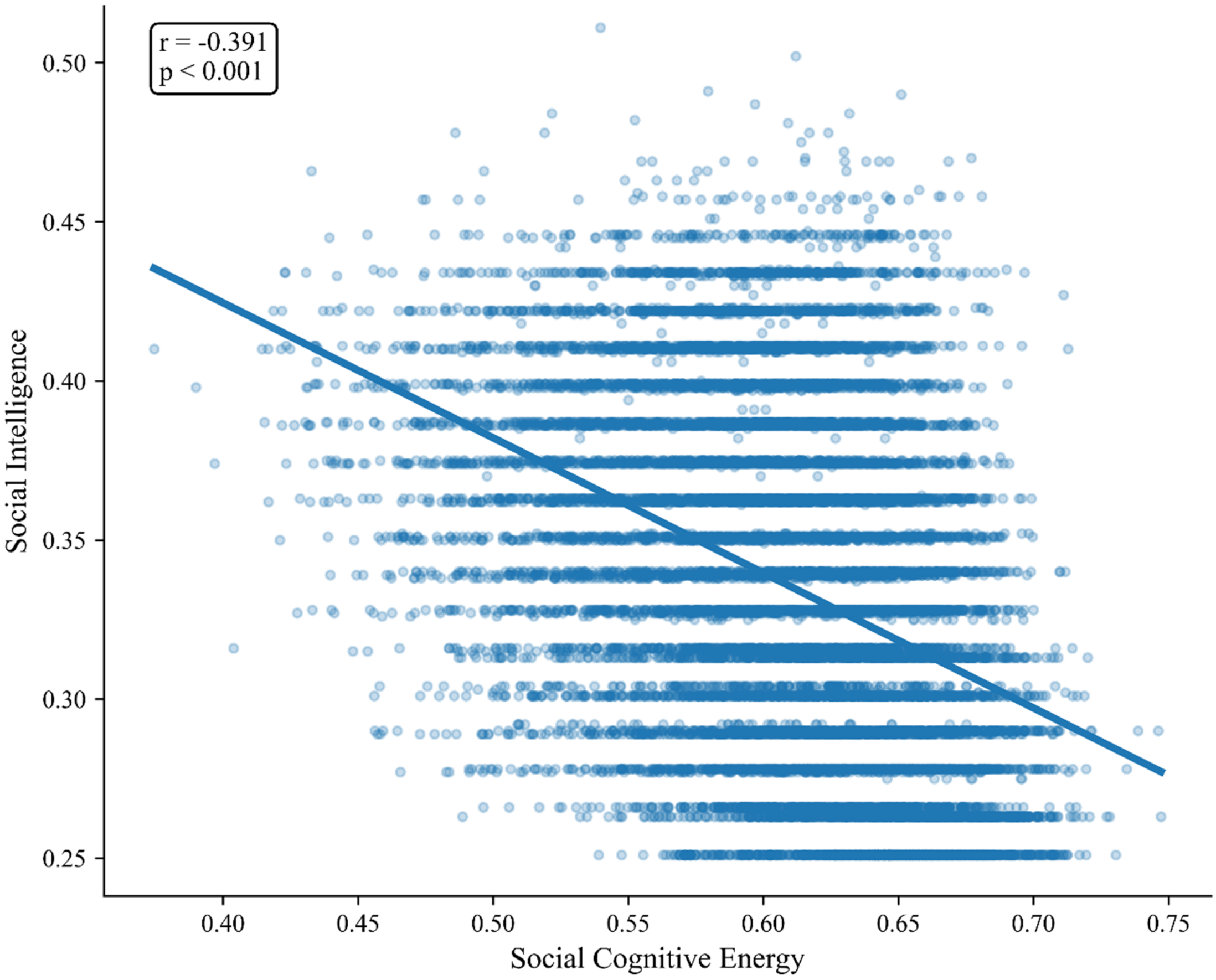

**Figure 10. Relationship between social cognitive energy and social intelligence.**

The analysis reveals a statistically significant negative association between social cognitive energy and social intelligence (($r=-0.391$, $p<0.001$)). Interaction states characterized by higher levels of social intelligence consistently exhibit lower energy values, whereas higher-energy states tend to be associated with less developed forms of social organization.

This result provides important empirical evidence for the energy optimization principle underlying HACD-H. Rather than requiring continuously increasing energy expenditure, social intelligence appears to emerge through the progressive reduction of social cognitive instability. As interactions become more adaptive, relationally coherent, and memory-consistent, the system moves toward increasingly stable regions of the energy landscape.

The observed negative relationship can be understood from the perspective of dynamical systems theory. High-energy states correspond to interaction configurations characterized by greater uncertainty, weaker relational organization, limited social memory integration, or reduced behavioral consistency. Such states require continual adjustment and therefore occupy relatively unstable positions within the social cognitive landscape. In contrast, low-energy states reflect interaction configurations in which multiple cognitive processes have become coordinated and mutually reinforcing.

This interpretation is consistent with the attractor structures and trust basins identified in previous analyses. Stable attractors correspond to local low-energy regions of the landscape, and interaction trajectories naturally evolve toward these regions through repeated adaptation. As trajectories approach low-energy basins, social intelligence increases because the system can devote fewer resources to maintaining coherence and more resources to effective social coordination.

Importantly, the results suggest that social intelligence should not be interpreted as a consequence of accumulating isolated cognitive capacities. Instead, intelligence emerges through the efficient organization of social cognitive processes. The ability to maintain coherent relationships, integrate interaction history, regulate emotional dynamics, and preserve personality consistency collectively reduces the energetic cost of adaptation while simultaneously increasing social competence.

The observed coupling therefore provides a mechanistic explanation for the emergence phenomenon reported in the previous section. Social intelligence develops as interaction trajectories progressively descend the social cognitive energy landscape toward more stable and efficient configurations. Higher intelligence is thus associated not with higher energy states, but with successful energy minimization and improved organizational efficiency.

Overall, the findings provide strong empirical support for the Energy–Intelligence Coupling hypothesis of HACD-H. The significant negative correlation demonstrates that social intelligence is closely linked to the energetic structure of the social cognitive landscape and supports the broader claim that long-term human–AI coevolution is governed by principles of energy optimization and dynamical self-organization.

## 4.8 Energy Optimization Dynamics

The preceding analyses demonstrated that social intelligence is negatively associated with social cognitive energy and that interaction trajectories tend to occupy increasingly stable regions of the reconstructed energy landscape. HACD-H further predicts that these observations are not isolated phenomena but manifestations of a continuous optimization process. As human–AI interactions evolve, the social cognitive system is expected to progressively reduce instability and move toward energetically favorable configurations.

To evaluate this prediction, we analyzed the long-term evolution of social cognitive energy across normalized interaction trajectories. Energy values were computed using the theory-guided HACD-H energy function, and temporal trends were reconstructed by aligning interaction trajectories according to their relative developmental progress.

Figure 11 presents the long-term dynamics of social cognitive energy.

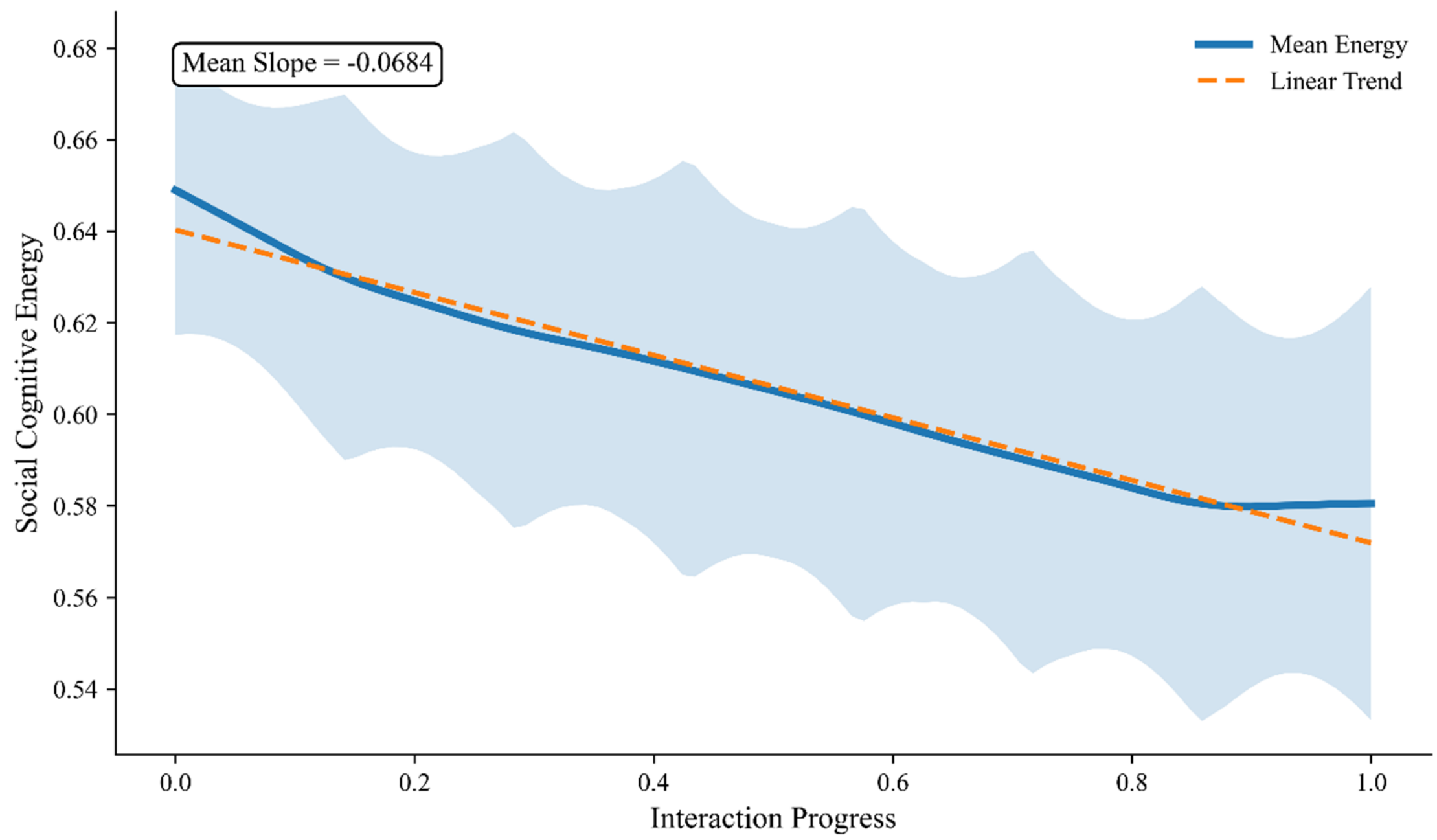

**Figure 11. Long-term energy optimization dynamics.**

The results reveal a persistent downward trend in social cognitive energy throughout the interaction process. The average trajectory exhibits a negative slope (Mean Slope = −0.0684), indicating that interaction states progressively move toward lower-energy regions of the social cognitive landscape as interaction history accumulates.

This finding provides direct empirical support for the energy optimization principle proposed by HACD-H. Rather than exhibiting unrestricted growth or random fluctuations, the social cognitive system demonstrates a systematic tendency toward energy reduction. Repeated interaction enables emotional regulation, relational stabilization, memory integration, and personality-consistent adaptation to become increasingly coordinated, thereby reducing the overall energetic cost required to maintain coherent social behavior.

The observed energy decline is consistent with the attractor structures identified in Figure 5 and the trust basins identified in Figure 6. Both analyses suggested the existence of stable regions capable of constraining trajectory evolution. The present results demonstrate that interaction trajectories do not merely occupy such regions temporarily but gradually converge toward them through long-term adaptation. This convergence process manifests as a progressive reduction in social cognitive energy.

The findings also provide a dynamical explanation for the emergence of social intelligence observed in Figure 8. As trajectories descend the energy landscape, the interaction system becomes increasingly organized and efficient. Emotional responses become more predictable, relational structures become more stable, and memory processes become more effectively integrated. These changes collectively contribute to the emergence of higher-order social intelligence while simultaneously reducing energetic instability.

Importantly, the optimization process observed here differs from conventional notions of performance maximization. HACD-H does not predict that successful social systems continuously accumulate energy. Instead, it predicts that adaptive systems reduce unnecessary cognitive and relational costs by organizing behavior into increasingly stable configurations. Social development therefore emerges through energy minimization rather than energy expansion.

From the perspective of complex adaptive systems, the observed dynamics indicate that long-term human–AI interaction exhibits self-organizing behavior. Local conversational adaptations accumulate across time and collectively drive the system toward globally stable regions of the state space. The resulting trajectory convergence is a hallmark of attractor-driven evolution and provides empirical evidence that human–AI coevolution is governed by large-scale organizational principles rather than isolated conversational events.

Overall, the results strongly support the Energy Optimization Dynamics hypothesis of HACD-H. Human–AI interaction trajectories exhibit a robust tendency toward lower-energy states, consistent with the reconstructed energy landscape, the observed attractor structures, and the negative coupling between energy and social intelligence. These findings suggest that long-term social development is fundamentally characterized by progressive movement toward stable and energetically efficient modes of interaction.

## 5. Discussion

### 5.1 Human–AI Interaction as a Self-Organizing Social Cognitive System

This study introduced HACD-H, a formal theory describing long-term human–AI interaction as a self-organizing social cognitive system. Unlike conventional approaches that treat emotion, memory, personality, or trust as isolated components, HACD-H conceptualizes social interaction as the coupled evolution of multiple social cognitive processes operating across different temporal scales.

The empirical findings provide consistent support for this perspective. Temporal persistence analysis revealed a clear hierarchy of characteristic timescales, demonstrating that emotional states fluctuate rapidly whereas personality-related structures remain comparatively stable. This organization suggests that human–AI interaction exhibits an intrinsic temporal architecture rather than random behavioral variation.

Furthermore, the identification of relational attractors and trust basins indicates that interaction trajectories do not evolve arbitrarily. Instead, repeated interaction gradually constrains behavioral dynamics and guides trajectories toward increasingly stable regions of the social cognitive state space. These findings are consistent with theories of self-organization in complex adaptive systems and suggest that long-term human–AI relationships exhibit emergent structural regularities.

## 5.2 Emergence of Social Intelligence Through Coevolution

One of the central claims of HACD-H is that social intelligence is not an independently programmed capability but an emergent property arising from long-term social cognitive coevolution.

The results strongly support this proposition. Developmental phase transition analysis revealed critical periods during which social intelligence increased rapidly, suggesting that interaction development is characterized by nonlinear organizational changes rather than smooth incremental growth. Similarly, the emergence patterns observed across interaction trajectories indicate that higher-order social capabilities arise through the coordinated interaction of emotional adaptation, relational organization, memory integration, and personality consistency.

These findings challenge the dominant view that social intelligence can be fully explained through increasingly powerful language models or larger training corpora. Instead, the results suggest that social intelligence is fundamentally a dynamical phenomenon emerging from the long-term organization of interaction processes.

From this perspective, social intelligence resembles other emergent properties observed in complex systems, where global organization arises from repeated local interactions without centralized control.

## 5.3 Energy Optimization as a Principle of Social Development

A particularly important contribution of the present study is the introduction and empirical validation of Social Cognitive Energy Theory.

The reconstructed energy landscape demonstrates that social interaction can be represented as movement through a structured state space characterized by energy gradients, attractors, and stable basins. More importantly, the empirical analyses revealed a significant negative relationship between social cognitive energy and social intelligence (($r=-0.391, p<0.001$)), indicating that more socially intelligent interaction states tend to occupy lower-energy regions of the landscape.

Longitudinal analysis further demonstrated a persistent decline in social cognitive energy across interaction trajectories. The negative average energy slope ((Mean\ Slope=- 0.0684)) suggests that repeated human–AI interaction is characterized by progressive energy optimization rather than energy accumulation.

This finding has important theoretical implications. Traditional developmental perspectives often emphasize growth, accumulation, or increasing complexity as the primary indicators of progress. In contrast, HACD-H suggests that successful social development involves the reduction of instability, uncertainty, and organizational cost. Social systems become more intelligent not because they continually acquire more energy, but because they organize existing resources more efficiently.

Consequently, long-term social development may be better understood as a process of convergence toward stable and energetically efficient configurations.

## 5.4 Implications for Socially Intelligent AI

The findings have direct implications for the design of next-generation socially intelligent AI systems. Most contemporary conversational agents remain optimized for short-term response quality and immediate task completion. Although such systems may generate fluent responses, they often lack stable long-term relational dynamics and coherent social development.

The present results suggest that socially intelligent AI should be designed as an adaptive dynamical system rather than a static language generator. Future systems may benefit from explicitly modeling emotional adaptation, relational memory, trust formation, and personality consistency as interacting components of a unified social cognitive architecture.

Furthermore, the concept of social cognitive energy provides a potential optimization objective for future AI systems. Instead of maximizing isolated conversational metrics, socially intelligent agents may seek to maintain interaction trajectories within low-energy and high-stability regions of the social cognitive landscape. Such an approach could promote more coherent, trustworthy, and sustainable long-term human–AI relationships.

## 5.5 Limitations and Future Directions

Several limitations should be acknowledged.

First, the empirical analyses were conducted using a reconstructed social cognitive dataset derived from a single conversational corpus. Although the dataset contains extensive interaction trajectories, future studies should evaluate HACD-H across multiple domains, languages, and interaction settings.

Second, several latent social cognitive variables were inferred through computational annotation models rather than directly observed. Improvements in affective computing, personality modeling, and social behavior measurement may further enhance the precision of state reconstruction.

Third, the present work focused primarily on dyadic human–AI interaction. Extending HACD-H to multi-agent environments, online communities, and large-scale social ecosystems represents an important direction for future research.

Finally, while the current study provides evidence supporting the proposed theoretical principles, future work should investigate whether the identified attractors, trust basins, and energy landscapes can be actively manipulated through intervention strategies designed to guide social development.

## 5.6 Toward a Theory of Human–AI Social Coevolution

The broader significance of HACD-H lies in its attempt to establish a unified theoretical foundation for long-term human–AI social interaction.

The results suggest that human–AI relationships are governed by identifiable organizational principles, including multi-timescale adaptation, attractor formation, trust basin development, social intelligence emergence, and energy optimization. Together, these mechanisms provide a coherent explanation for how stable social relationships can arise through repeated interaction.

More generally, HACD-H shifts the focus of social AI research from isolated conversational behaviors toward the dynamics of long-term social development. Understanding interaction as a process of coevolution may provide a foundation for future theories of socially adaptive artificial intelligence and contribute to the development of AI systems capable of sustaining meaningful and coherent relationships over extended periods of time.

# 6. Conclusion

This study introduced the Human–AI Coevolution Dynamics Framework (HACD-H), a formal theory for understanding long-term human–AI interaction as a self-organizing social cognitive system. Unlike existing approaches that primarily focus on isolated mechanisms such as emotion modeling, memory retrieval, or personality conditioning, HACD-H provides a unified dynamical perspective in which emotional adaptation, relational organization, social memory, and personality consistency jointly shape the evolution of interaction trajectories across time.

The proposed framework conceptualizes human–AI interaction as the evolution of latent social cognitive states operating on multiple temporal scales. Building upon this perspective, HACD-H introduces a set of theoretical principles describing temporal persistence hierarchies, relational attractor formation, trust basin development, developmental phase transitions, social cognitive energy landscapes, social intelligence emergence, energy–intelligence coupling, and long-term energy optimization.

To evaluate these propositions, we constructed a socially enriched conversational dataset containing approximately 14,700 interaction turns and developed a theory-driven empirical validation framework. The results provided consistent evidence supporting the major predictions of HACD-H. Temporal persistence analysis revealed a clear hierarchy of social cognitive timescales. State-space reconstruction identified stable relational attractors and trust basins. Developmental analyses demonstrated phase-transition-like changes in social intelligence trajectories. Energy landscape reconstruction revealed structured organizational patterns within the social cognitive state space. Furthermore, social intelligence exhibited a significant negative association with social cognitive energy, while long-term interaction trajectories displayed progressive energy optimization characterized by decreasing energy over time.

Taken together, these findings suggest that socially intelligent human–AI relationships emerge through a process of long-term social cognitive coevolution. Rather than accumulating isolated capabilities, interaction systems gradually reorganize themselves into increasingly stable, coherent, and energetically efficient configurations. From this perspective, social intelligence is best understood as an emergent property arising from the coordinated dynamics of multiple interacting social cognitive processes.

The theoretical implications extend beyond conversational AI. HACD-H provides a general framework for studying adaptive social systems and offers a bridge between artificial intelligence, cognitive science, affective computing, and complex systems research. By introducing concepts such as relational attractors, trust basins, and social cognitive energy landscapes, the framework establishes a new vocabulary for describing the dynamics of long-term social interaction.

The practical implications are equally significant. Current conversational agents are typically optimized for short-term response quality, whereas the present findings suggest that future socially intelligent systems should be designed to support long-term adaptation, relationship development, trust formation, and energy-efficient social coordination. Such systems may ultimately be capable of maintaining coherent and meaningful social relationships over extended periods of interaction.

Future research may extend HACD-H to multimodal interaction, multi-agent environments, online social ecosystems, and embodied AI systems. Further investigation of intervention strategies capable of shaping attractor structures, trust basins, and energy landscapes may also provide new approaches for guiding the development of socially adaptive artificial intelligence.

Overall, HACD-H represents an initial step toward a general theory of human–AI social coevolution. The framework suggests that long-term interaction is governed by identifiable organizational principles and that social intelligence emerges through the self-organization of social cognitive dynamics across time. Understanding these principles may contribute to the development of the next generation of socially intelligent AI systems and advance the scientific study of human–AI relationships.

## Data Availability Statement

The full dataset, including raw conversational trajectories and processed social cognitive annotations, is publicly available on ScienceDB (https://www.scidb.cn/s/2emEZr) to support reproducibility and further research.

## REFERENCES

Brinkschulte, L., Schlögl, S., Monz, A., Schöttle, P., & Janetschek, M. (2022). Perspectives on socially intelligent conversational agents. *Multimodal Technologies and Interaction, 6*(8), Article 62. https://doi.org/10.3390/mti6080062.

Brown, T. B., Mann, B., Ryder, N., Subbiah, M., Kaplan, J. D., Dhariwal, P., Neelakantan, A., Shyam, P., Sastry, G., Askell, A., Agarwal, S., Herbert-Voss, A., Krueger, G., Henighan, T., Child, R., Ramesh, A., Ziegler, D., Wu, J., Winter, C., ... & Amodei, D. (2020). Language models are few-shot learners. *Advances in Neural Information Processing Systems, 33*, 1877–1901.

Burgoon, J. K., Stern, L. A., & Dillman, L. (1995). *Interpersonal adaptation: Dyadic interaction patterns*. Cambridge University Press. https://doi.org/10.1017/CBO9780511623668.

Cai, M., Wang, D., Feng, S., & Zhang, Y. (2024). PECER: Empathetic response generation via dynamic personality extraction and contextual emotional reasoning. In *ICASSP 2024-2024 IEEE International Conference on Acoustics, Speech and Signal Processing (ICASSP)* (pp. 10631–10635). IEEE. https://doi.org/10.1109/ICASSP48485.2024.10448008.

Chen, A., Du, C., Chen, J., Xu, J., Zhang, Y., Yuan, S., Li, J., Wang, H., Liu, Y., Li, B., Xu, H., Xu, B., Li, Y., Zhang, Y., Wang, J., Li, Y., Chen, Y., Zhang, J., Li, Y., ... & Xiao, Y. (2025). Deeper insight into your user: Directed persona refinement for dynamic persona modeling. In *Proceedings of the 63rd Annual Meeting of the Association for Computational Linguistics (Volume 1: Long Papers)* (pp. 24157–24180). Association for Computational Linguistics. https://doi.org/10.18653/v1/2025.acl-long.1293.

Guo, S., & Ning, B. (2024). A review of empathetic conversational systems. In *2024 9th International Conference on Intelligent Computing and Signal Processing (ICSP)* (pp. 1279–1286). IEEE. https://doi.org/10.1109/ICSP62051.2024.10881719.

Gweon, H., Fan, J., & Kim, B. (2023). Socially intelligent machines that learn from humans and help humans learn. *Philosophical Transactions of the Royal Society A*, *381*(2251), 20220048.

Han, X., Zhao, W., Guan, Z., & Peng, J. (2025). Act-LLM: A whole-process chain for character-centric role-playing with large language models. *Expert Systems with Applications*, 129024.

Hong, S., Zhuge, M., Chen, J., Zheng, X., Cheng, Y., Wang, J., ... & Schmidhuber, J. (2024, May). MetaGPT: Meta programming for a multi-agent collaborative framework. In *International Conference on Learning Representations* (Vol. 2024, pp. 23247-23275).

Hui, B., Yang, J., Cui, Z., Yang, J., Liu, D., Zhang, L., Tian, L., Zhang, Q., Yu, B., Lu, K., Li, J., Dong, Y., Xue, F., Ren, J., Ren, X., & Lin, J. (2024). Qwen2.5-coder technical report. arXiv preprint. https://arxiv.org/abs/2409.12186.

Izard, C. E. (1993). Four systems for emotion activation: Cognitive and noncognitive processes. *Psychological Review, 100*(1), 68–90. https://doi.org/10.1037/0033-295X.100.1.68.

Järvelä, S., Zhao, G., Nguyen, A., & Chen, H. (2025). Hybrid intelligence: Human–AI coevolution and learning. *British Journal of Educational Technology*, *56*(2), 455.

Khalil, H. K. (1996). Adaptive output feedback control of nonlinear systems represented by input-output models. *IEEE Transactions on Automatic Control, 41*(2), 177–188. https://doi.org/10.1109/9.481517.

Kirk, H. R., Gabriel, I., Summerfield, C., Vidgen, B., & Hale, S. A. (2025). Why human–AI relationships need socioaffective alignment. *Humanities and Social Sciences Communications*, *12*(1), 1-9.

Kosinski, M. (2024). Evaluating large language models in theory of mind tasks. *Proceedings of the National Academy of Sciences*, *121*(45), e2405460121.

Kruglanski, A. W., Jasko, K., Milyavsky, M., Chernikova, M., Webber, D., Pierro, A., & Di Santo, D. (2018). Cognitive consistency theory in social psychology: A paradigm reconsidered. *Psychological Inquiry, 29*(2), 45–59. https://doi.org/10.1080/1047840X.2018.1480613.

Li, G., Hammoud, H., Itani, H., Khizbullin, D., & Ghanem, B. (2023). Camel: Communicative agents for" mind" exploration of large language model society. *Advances in neural information processing systems*, *36*, 51991-52008.

Liu, Q., Chen, Y., Chen, B., Lou, J. G., Chen, Z., Zhou, B., & Zhang, D. (2020). You impress me: Dialogue generation via mutual persona perception. In *Proceedings of the 58th Annual Meeting of the Association for Computational Linguistics* (pp. 1417–1427). Association for Computational Linguistics. https://doi.org/10.18653/v1/2020.acl-main.131.

Lu, J., Li, D., Ding, B., & Kang, Y. (2024). Improving embedding with contrastive fine-tuning on small datasets with expert-augmented scores. *arXiv preprint arXiv:2408.11868*.

Luszczynska, A., & Schwarzer, R. (2005). Social cognitive theory. In M. Conner & P. Norman (Eds.), *Predicting health behaviour* (2nd ed., pp. 127–169). Open University Press.

Lyapunov, A. M. (1992). The general problem of the stability of motion. *International Journal of Control, 55*(3), 531–534. https://doi.org/10.1080/00207179208934253 (Original work published 1892).

Majumder, N., Hong, P., Peng, S., Lu, J., Ghosal, D., Gelbukh, A., Akhtar, M. S., & Poria, S. (2020). MIME: MIMicking emotions for empathetic response generation. In *Proceedings of the 2020 Conference on Empirical Methods in Natural Language Processing (EMNLP)* (pp. 8968–8979). Association for Computational Linguistics. https://doi.org/10.18653/v1/2020.emnlp-main.721.

Mazaré, P. E., Humeau, S., Raison, M., & Bordes, A. (2018). Training millions of personalized dialogue agents. In *Proceedings of the 2018 conference on empirical methods in natural language processing* (pp. 2775-2779).

Milnor, J. (1985). On the concept of attractor. *Communications in Mathematical Physics, 99*(2), 177–195. https://doi.org/10.1007/BF01212280.

Noller, J. (2025). 4E cognition and the coevolution of human–AI interaction. *Discover Artificial Intelligence*, *5*(1), 323.

Ouyang, L., Wu, J., Jiang, X., Almeida, D., Wainwright, C., Mishkin, P., Zhang, C., Agarwal, S., Slama, K., Ray, A., Schulman, J., Hilton, J., Kelton, F., Miller, L., Simens, M., Askell, A., Welinder, P., Christiano, P. F., Leike, J., & Lowe, R. (2022). Training language models to follow instructions with human feedback. *Advances in Neural Information Processing Systems, 35*, 27730–27744.

Packer, C., Fang, V., Patil, S. G., Lin, K., Wooders, S., & Gonzalez, J. E. (2023). *MemGPT: Towards LLMs as operating systems*. arXiv preprint. https://arxiv.org/abs/2310.08560.

Park, J. S., O'Brien, J., Cai, C. J., Morris, M. R., Liang, P., & Bernstein, M. S. (2023, October). Generative agents: Interactive simulacra of human behavior. In *Proceedings of the 36th annual acm symposium on user interface software and technology* (pp. 1-22).

Pedreschi, D., Pappalardo, L., Ferragina, E., Baeza-Yates, R., Barabási, A. L., Dignum, F., ... & Vespignani, A. (2025). Human-AI coevolution. *Artificial Intelligence*, *339*, 104244.

Qian, Y., & Wan, X. (2025). Co-experiencing with AI: Effects on social bonding and empathy in human-AI relationships. *Technology in society*, *83*, 103009.

Rainey, P. B., & Hochberg, M. E. (2025). Could humans and AI become a new evolutionary individual?. *Proceedings of the National Academy of Sciences*, *122*(37), e2509122122.

Rashkin, H., Smith, E. M., Li, M., & Boureau, Y. L. (2019). Towards empathetic open-domain conversation models: A new benchmark and dataset. In *Proceedings of the 57th Annual Meeting of the Association for Computational Linguistics* (pp. 5370–5381). Association for Computational Linguistics. https://doi.org/10.18653/v1/P19-1534

Shanahan, M., McDonell, K., & Reynolds, L. (2023). Role play with large language models. *Nature*, *623*(7987), 493-498.

Shao, Y., Li, L., Dai, J., & Qiu, X. (2023, December). Character-llm: A trainable agent for role-playing. In *Proceedings of the 2023 Conference on Empirical Methods in Natural Language Processing* (pp. 13153-13187).

Shuster, K., Xu, J., Komeili, M., Ju, D., Smith, E. M., Roller, S., Ung, M., Chen, M., Arora, K., Lane, J., BehnamGhader, P., Ning, S., Kambadur, M., & Weston, J. (2022). *BlenderBot 3: A deployed conversational agent that continually learns to responsibly engage*. arXiv preprint. https://arxiv.org/abs/2208.03188.

Strogatz, S. H. (2024). *Nonlinear dynamics and chaos: With applications to physics, biology, chemistry, and engineering* (3rd ed.). Chapman and Hall/CRC. https://doi.org/10.1201/9780429492563.

Vaswani, A., Shazeer, N., Parmar, N., Uszkoreit, J., Jones, L., Gomez, A. N., Kaiser, Ł., & Polosukhin, I. (2017). Attention is all you need. *Advances in Neural Information Processing Systems, 30*, 5998–6008. https://doi.org/10.48550/arXiv.1706.03762.

Wang, N., Peng, Z., Que, H., Liu, J., Zhou, W., Wu, Y., ... & Peng, J. (2024, August). Rolellm: Benchmarking, eliciting, and enhancing role-playing abilities of large language models. In *Findings of the Association for Computational Linguistics: ACL 2024* (pp. 14743-14777).

Wen, Z., Cao, J., Shen, J., Yang, R., Liu, S., & Sun, M. (2024). Personality-affected emotion generation in dialog systems. *ACM Transactions on Information Systems, 42*(5), Article 117. https://doi.org/10.1145/3641244.

Weston, J., Chopra, S., & Bordes, A. (2015). Memory networks. In *Proceedings of the 32nd International Conference on Machine Learning (ICML 2015)* (pp. 938–946). JMLR.org. http://proceedings.mlr.press/v37/weston15.html

Wu, Q., Bansal, G., Zhang, J., Wu, Y., Li, B., Zhu, E., ... & Wang, C. (2024, August). Autogen: Enabling next-gen LLM applications via multi-agent conversations. In *First conference on language modeling*.

Xu, X., Gou, Z., Wu, W., Niu, Z. Y., Wu, H., Wang, H., & Wang, S. (2022). Long time no see! Open-domain conversation with long-term persona memory. *In Findings of the Association for Computational Linguistics: ACL 2022 (pp. 2639–2650). Association for Computational Linguistics*. https://doi.org/10.18653/v1/2022.findings-acl.207.

Zhang, S., Dinan, E., Urbanek, J., Szlam, A., Kiela, D., & Weston, J. (2018). Personalizing dialogue agents: I have a dog, do you have pets too? In *Proceedings of the 56th Annual Meeting of the Association for Computational Linguistics (Volume 1: Long Papers)* (pp. 2204–2213). Association for Computational Linguistics. https://doi.org/10.18653/v1/P18-1205.

Zhong, W., Guo, L., Gao, Q., Ye, H., & Wang, Y. (2024, March). Memorybank: Enhancing large language models with long-term memory. In *Proceedings of the AAAI conference on artificial intelligence* (Vol. 38, No. 17, pp. 19724-19731).

Zhou, J., Luo, S., & Chen, H. (2025). Expansion quantization network: A micro-emotion detection and annotation framework. *PLOS ONE, 20(11),* Article e0333930. https://doi.org/10.1371/journal.pone.0333930.

Zhou, J., Luo, S., & Chen, H. (2025). H3P: High-performance personality prediction system based on large language model. *In 2025 IEEE 8th International Conference on Computer and Communication Engineering Technology (CCET)* (pp. 196–200). IEEE. https://doi.org/10.1109/CCET64841.2025.00107.

.